\documentclass[runningheads]{llncs}
\usepackage{graphicx}

\begin{document}

\title{VIENA$^2$: A Driving Anticipation Dataset} 
\titlerunning{VIENA$^2$: A Driving Anticipation Dataset} 



\author{Mohammad Sadegh Aliakbarian\inst{1,2,4} \and
Fatemeh Sadat Saleh\inst{1,4} \and
Mathieu Salzmann\inst{3} \and
Basura Fernando\inst{2} \and
Lars Petersson \inst{1,4} \and
Lars Andersson \inst{4}
}

\index{Aliakbarian, Mohammad Sadegh}
\index{Saleh, Fatemeh Sadat}

\authorrunning{M. S. Aliakbarian et al.} 


\institute{$^1$ANU, $^2$ACRV, $^3$CVLab, EPFL, $^4$Data61-CSIRO\\
\email{\{fname.lname\}@data61.csiro.au, mathieu.salzmann@epfl.ch, basura.fernando@anu.edu.au}}

\maketitle

\begin{abstract}
Action anticipation is critical in scenarios where one needs to react before the action is finalized. This is, for instance, the case in automated driving, where a car needs to, e.g., avoid hitting pedestrians and respect traffic lights. While solutions have been proposed to tackle subsets of the driving anticipation tasks, by making use of diverse, task-specific sensors, there is no single dataset or framework that addresses them all in a consistent manner. In this paper, we therefore introduce a new, large-scale dataset, called VIENA$^2$, covering 5 generic driving scenarios, with a total of 25 distinct action classes. It contains more than 15K full HD, 5s long videos acquired in various driving conditions, weathers, daytimes and environments, complemented with a common and realistic set of sensor measurements. This amounts to more than 2.25M frames, each annotated with an action label, corresponding to 600 samples per action class. We discuss our data acquisition strategy and the statistics of our dataset, and benchmark state-of-the-art action anticipation techniques, including a new multi-modal LSTM architecture with an effective loss function for action anticipation in driving scenarios.

\end{abstract}
\section{Introduction}
\label{sec:introduction}
Understanding actions/events from videos is key to the success of many real-world applications, such as autonomous navigation, surveillance and sports analysis. While great progress has been made to recognize actions from complete sequences~\cite{feichtenhofer2016convolutional,LRCN,TSN,DynamicNetwork}, action anticipation, which aims to predict the observed action as early as possible, has only reached a much lesser degree of maturity~\cite{aliakbarian2017encouraging,vondrick2016anticipating,soomro2016predicting}. Nevertheless, anticipation is a crucial component in scenarios where a system needs to react quickly, such as in robotics~\cite{koppula2016anticipating}, and automated driving~\cite{jain2016brain4cars,liebner2013generic,li2017unified}. Its benefits have also been demonstrated in surveillance settings~\cite{ramanathan2016detecting,wang2017hierarchical}.

\begin{figure}[t]
\centering
\begin{tabular}{c c c c c}
\multicolumn{5}{c}{\includegraphics[width=.98\textwidth]{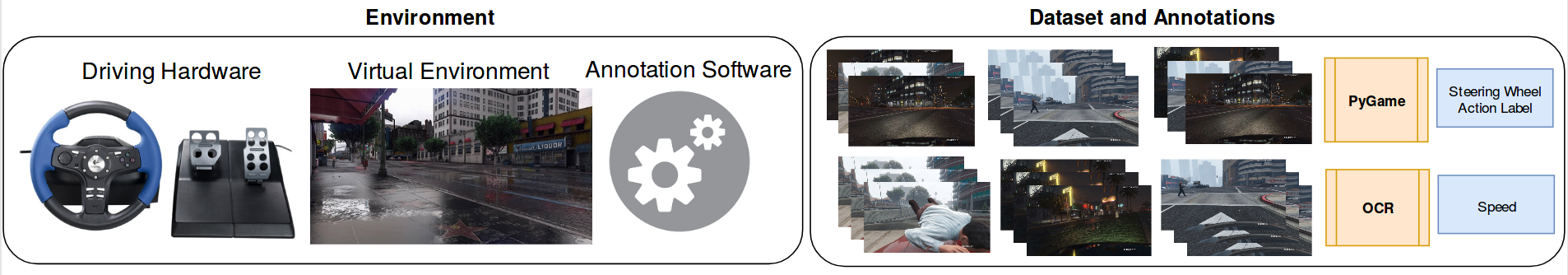}}\\
\includegraphics[width=.189\textwidth]{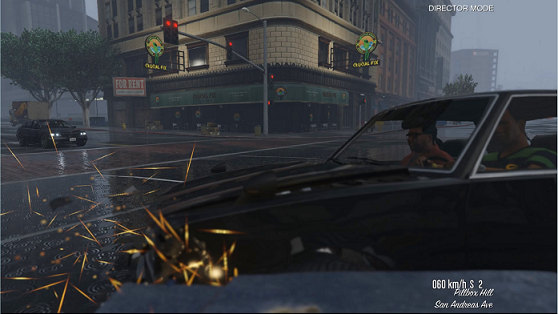} &
\includegraphics[width=.189\textwidth]{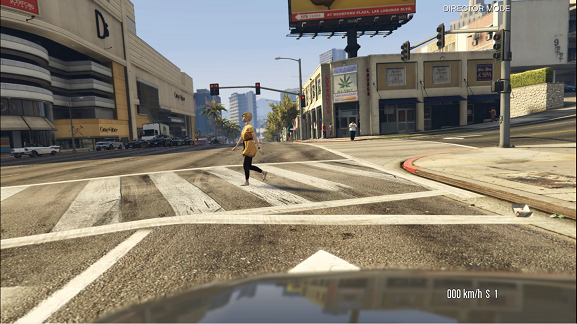} &
\includegraphics[width=.189\textwidth]{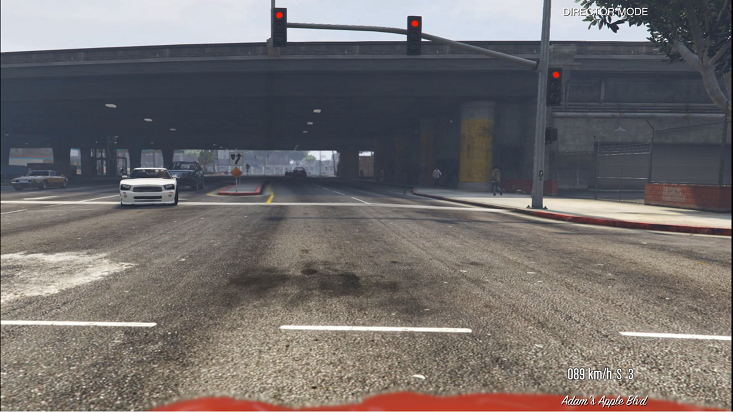} &
\includegraphics[width=.189\textwidth]{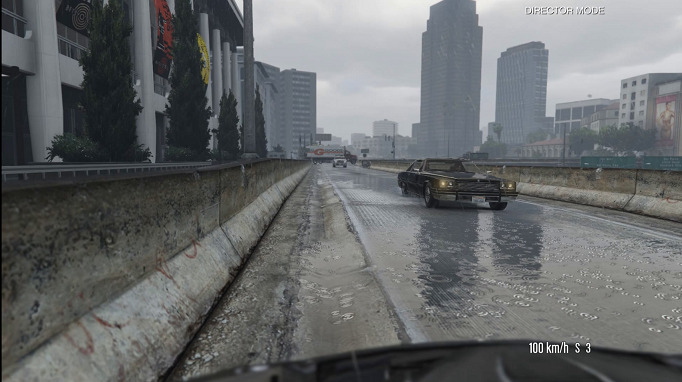} & 
\includegraphics[width=.189\textwidth]{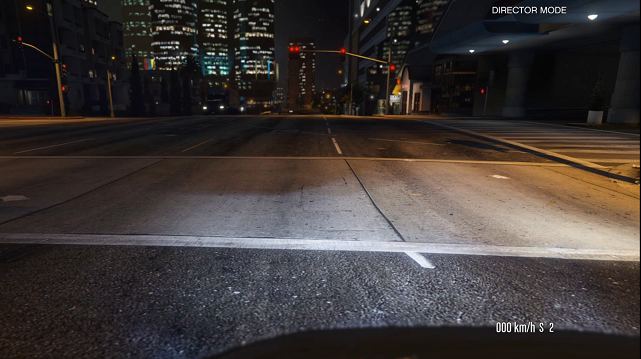} \\
\end{tabular}
\caption{{\bf Overview of our data collection.} Using the GTA V environment and driving equipment depicted in the top left box, we captured a new dataset covering 5 generic scenarios, illustrated in the right box, each containing multiple action classes (samples in bottom row). For more examples and examples of the vehicles our data was gathered with, please check our supplementary material.}
\label{fig:Viena2Overal}
\end{figure}

In this paper, we focus on the driving scenario. In this context, when consulting the main actors in the field, may they be from the computer vision community, the intelligent vehicle one or the automotive industry, the consensus is that predicting the intentions of a car's own driver, for Advanced Driver Assistance Systems (ADAS), remains a challenging task for a computer, despite being relatively easy for a human~\cite{dong2017intention,olabiyi2017driver,jain2016recurrent,jain2016brain4cars,rasouli2017agreeing}. Anticipation then becomes even more complex when one considers the maneuvers of other vehicles and pedestrians~\cite{klingelschmitt2016probabilistic,zyner2017long,dong2017intention}. However, it is key to avoiding dangerous situations, and thus to the success of autonomous driving.

Over the years, the researchers in the field of anticipation for driving scenarios have focused on specific subproblems of this challenging task, such as lane change detection~\cite{morris2011lane,tawari2014looking}, a car's own driver's intention~\cite{ohn2014head} or maneuver recognition~\cite{jain2015car,jain2016recurrent,jain2016brain4cars,olabiyi2017driver} and pedestrian intention prediction
~\cite{rasouli2017agreeing,pool2017using,li2017unified,schulz2015controlled}. 
Furthermore, these different subproblems are typically addressed by making use of different kinds of sensors, without considering the fact that, in practice, the automotive industry might not be able/willing to incorporate all these different sensors to address all these different tasks.

In this paper, we study the general problem of anticipation in driving scenarios, 
encompassing all the subproblems discussed above, and others, such as other drivers' intention prediction, with a fixed, sensible set of sensors. To this end, we introduce the \textbf{VI}rtual \textbf{EN}vironment for \textbf{A}ction \textbf{A}nalysis (VIENA$^2$) dataset, covering the five different subproblems of predicting driver maneuvers, pedestrian intentions, front car intentions, traffic rule violations, and accidents.
Altogether, these subproblems encompass a total of 25 distinct action classes. 
VIENA$^2$ was acquired using the GTA V video game~\cite{GTAgame}. It contains more than 15K full HD, 5s long videos, corresponding to more than 600 samples per action class, acquired in various driving conditions, weathers, daytimes, and environments. This amounts to more than 2.25M frames, each annotated with an action label. These videos are complemented by basic vehicle dynamics measurements, reflecting well the type of information that one could have access to in practice.

Below, we describe how VIENA$^2$ was collected and compare its statistics and properties to existing datasets. We then benchmark state-of-the-art action anticipation algorithms on VIENA$^2$, and introduce a new multi-modal, LSTM-based architecture, together with a new anticipation loss, which outperforms existing approaches in our driving anticipation scenarios. Finally, we investigate the benefits of our synthetic data to address anticipation from real images.
In short, our contributions are: {\bf (i)} a large-scale action anticipation dataset for general driving scenarios; {\bf(ii)} a multi-modal action anticipation architecture. 

VIENA$^2$ is meant as an extensible dataset that will grow over time to include not only more data but also additional scenarios. Note that, for benchmarking purposes, however, we will clearly define training/test partitions. A similar strategy was followed by other datasets such as CityScapes, which contains a standard benchmark set but also a large amount of additional data. VIENA$^2$ is publicly available, together with our benchmark evaluation, our new architecture and our multi-domain training strategy.

\section{VIENA$^2$}
\label{sec:viena2}
VIENA$^2$ is a large-scale dataset for action anticipation, and more generally action analysis, in driving scenarios.
While it is generally acknowledged that anticipation is key to the success of automated driving, 
to the best of our knowledge, there is currently no dataset that covers a wide range of scenarios with a common, yet sensible set of sensors. Existing datasets focus on specific subproblems, such as driver maneuvers
and pedestrian intentions~\cite{rasouli2017agreeing,pool2017using,kooij2014context}, and make use of different kinds of sensors. Furthermore, with the exception of~\cite{jain2016brain4cars}, none of these datasets provide videos whose first few frames do not already show the action itself or the preparation of the action. 
To create VIENA$^2$, we made use of the GTA V video game, whose publisher allows, under some conditions, for the non-commercial use of the footage~\cite{GTA_file}. Beyond the fact that, as shown in~\cite{richterplaying} via psychophysics experiments, GTA V provides realistic images that can be captured in varying weather and daytime conditions, it has the additional benefit of allowing us to cover crucial anticipation scenarios, such as accidents, for which real-world data would be virtually impossible to collect. 
In this section, we first introduce the different scenarios covered by VIENA$^2$ and discuss the data collection process. We then study the statistics of VIENA$^2$ and compare it against existing datasets.

\subsection{Scenarios and Data Collection}
\label{sec:scenarios}

\begin{figure*}[t]
\centering
\scriptsize
\begin{tabular}{|c |c |c|}
\hline
Driver Maneuver & Accident & Traffic Rule\\
\includegraphics[width=0.3\textwidth]{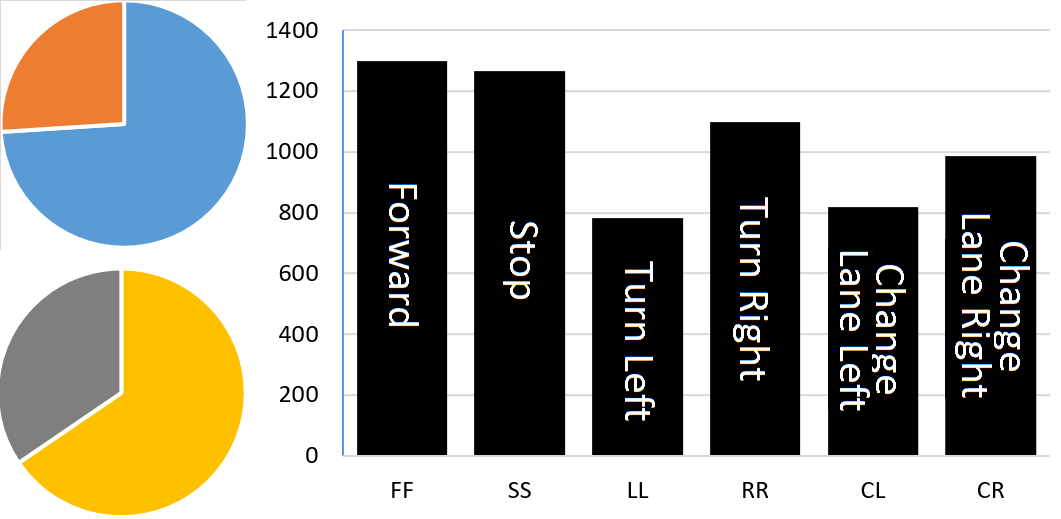} & 
\includegraphics[width=0.3\textwidth]{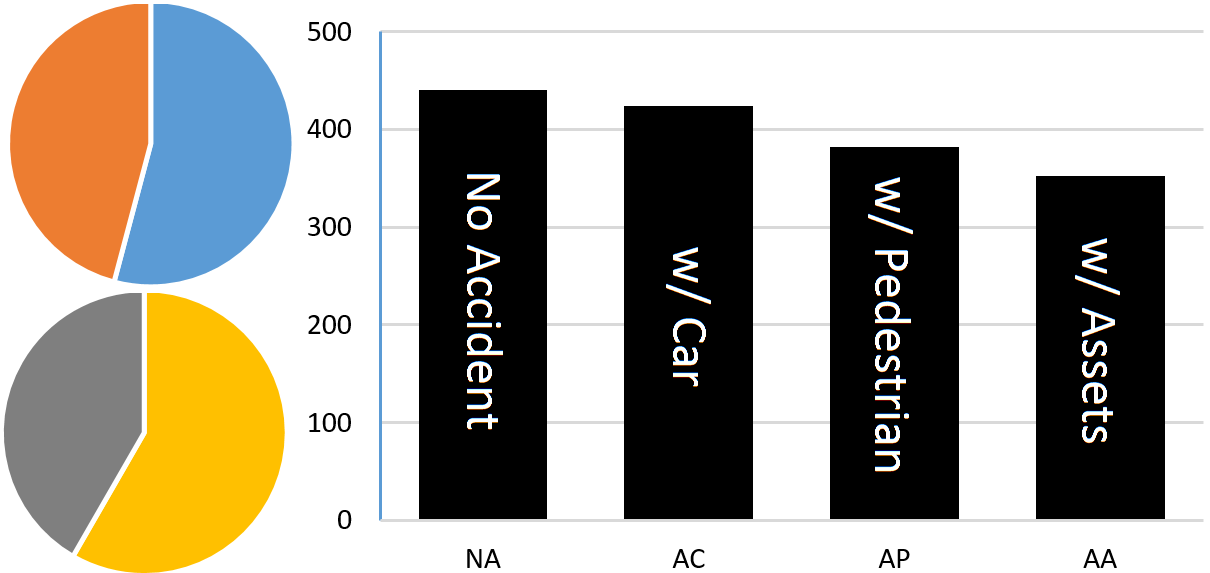} & 
\includegraphics[width=0.3\textwidth]{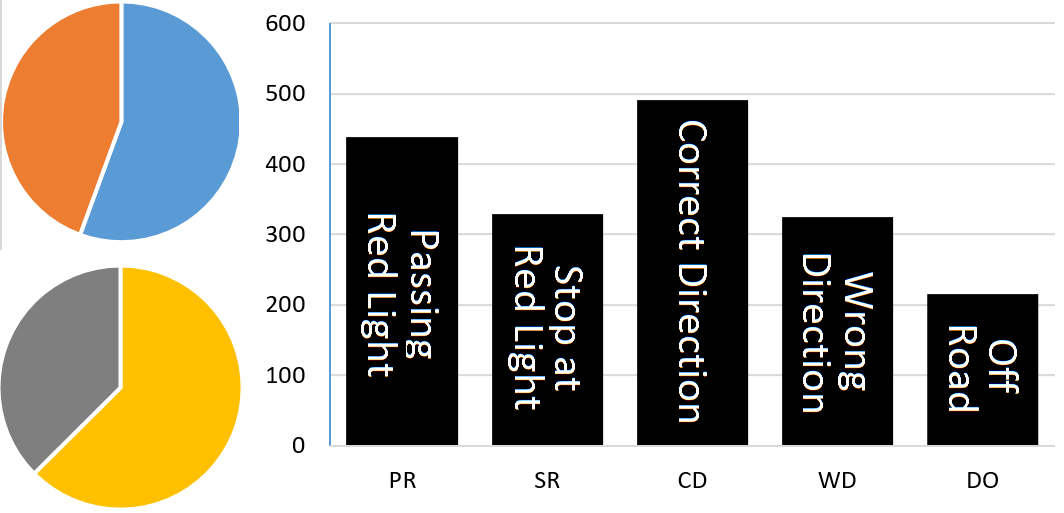} \\
\hline
Pedestrian Intention & Front Car Intention & Heavy Vehicle Maneuver\\ 
\includegraphics[width=0.3\textwidth]{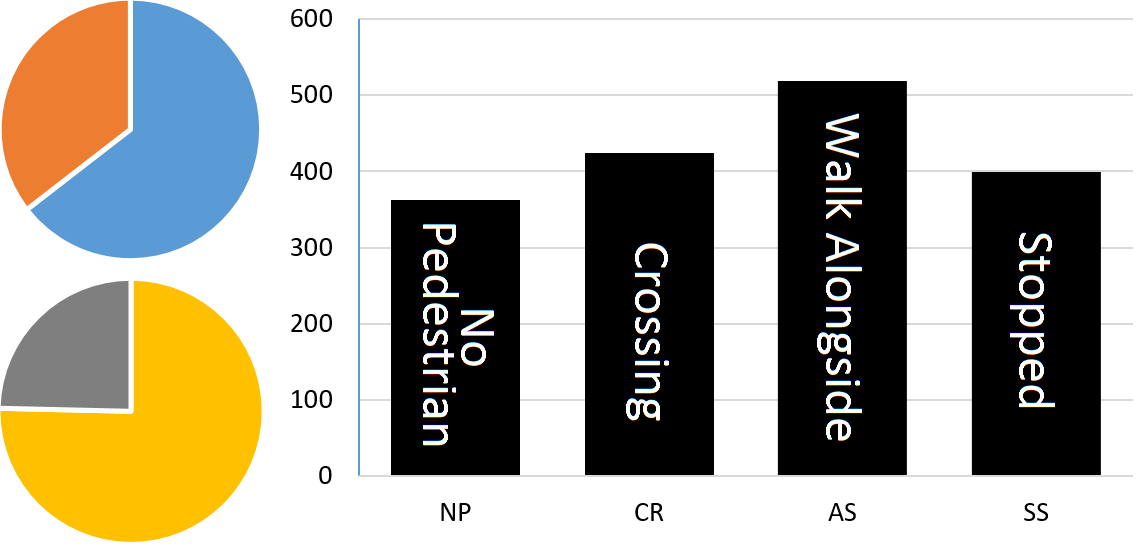}&\includegraphics[width=0.3\textwidth]{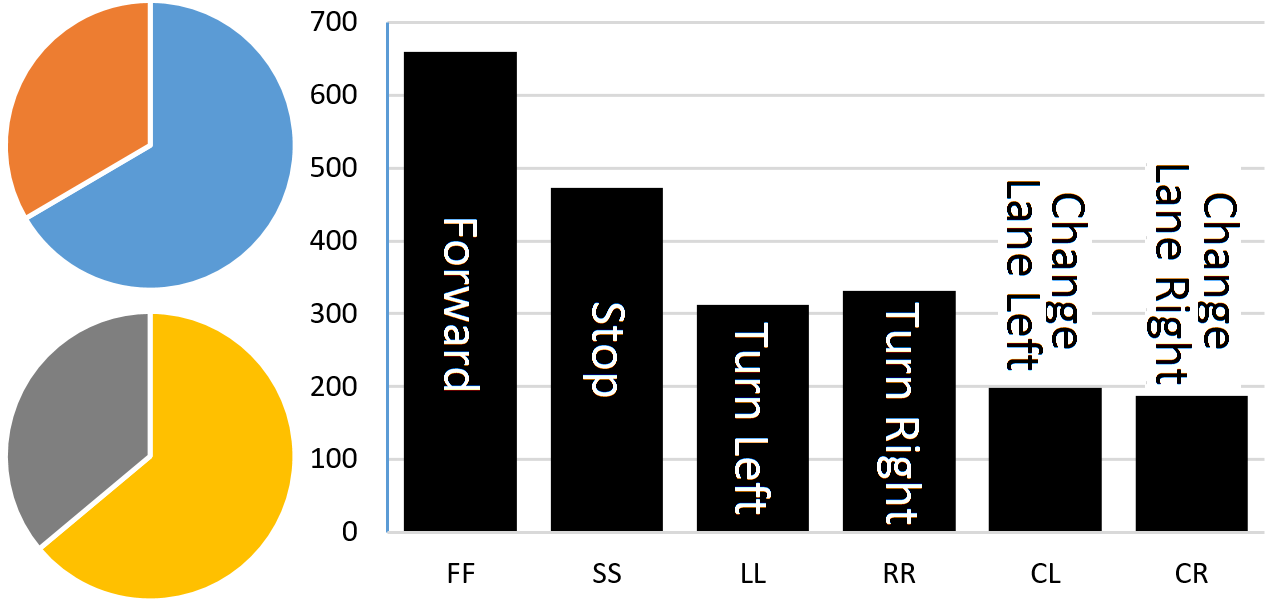} & 
\includegraphics[width=0.3\textwidth]{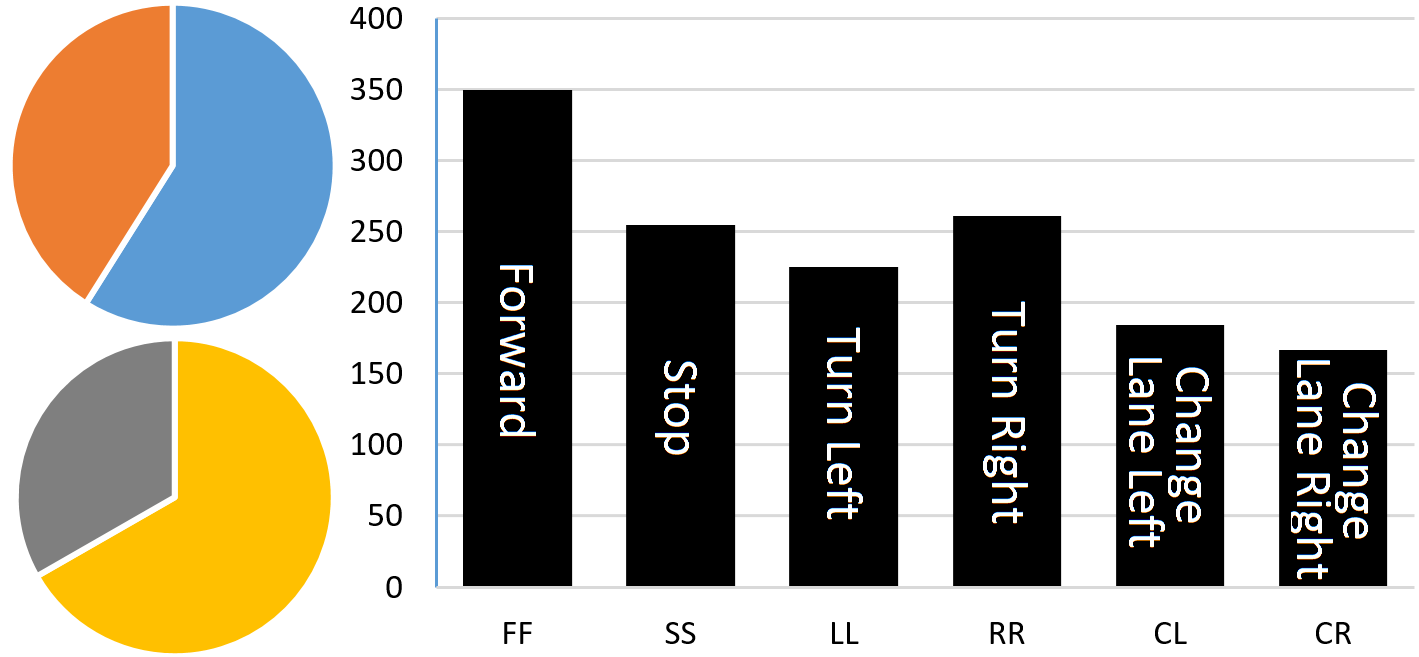}\\
\hline
\end{tabular}
\caption{{\bf Statistics for each scenario of VIENA$^2$.} We plot the number of videos per class, and proportions of different weather conditions (clear in yellow vs rainy/snowy in gray) and different daytime (day in orange vs night in blue). Best seen in color.}
\label{fig:plot_stat}
\end{figure*}

As illustrated in Fig.~\ref{fig:plot_stat}, VIENA$^2$ covers five generic driving scenarios.
These scenarios are all human-centric, i.e., consider the intentions of humans, but three of them focus on the car's own driver, while the other two relate to the environment (i.e., pedestrians and other cars). These scenarios are:
\begin{enumerate}
\item \textbf{Driver Maneuvers (DM).} This scenario covers the 6 most common maneuvers a driver performs while driving: Moving forward (FF), stopping (SS), turning (left (LL) and right (RR)) and changing lane (left (CL) and right (CR)). Anticipation of such maneuvers as early as possible is critical in an ADAS context to avoid dangerous situations.

\item \textbf{Traffic Rules (TR).} This scenario contains sequences depicting the car's own driver either violating or respecting traffic rules, e.g., stopping at (SR) and passing (PR) a red light, driving in the (in)correct direction (WD,CD), and driving off-road (DO). Forecasting these actions is also crucial for ADAS.

\item \textbf{Accidents (AC).} In this scenario, we capture the most common real-world accident cases: Accidents with other cars (AC), with pedestrians (AP), and with assets (AA), such as buildings, traffic signs, light poles and benches, as well as no accident (NA).
Acquiring such data in the real world is virtually infeasible. Nevertheless, these actions are crucial to anticipate for ADAS and autonomous driving.

\item \textbf{Pedestrian Intentions (PI).} This scenario addresses the question of whether a pedestrian is going to cross the road (CR), or has stopped (SS) but does not want to cross, or is walking along the road (AS) (on the sidewalk). We also consider the case where no pedestrian is in the scene (NP). As acknowledged in the literature~\cite{pool2017using,schulz2015controlled,rasouli2017agreeing}, early understanding of pedestrians' intentions is critical for automated driving.

\item \textbf{Front Car Intentions (FCI).} The last generic scenario of VIENA$^2$ aims at anticipating the maneuvers of the front car. This knowledge has a strong influence on the behavior to adopt to guarantee safety. The classes are same as the ones in Driver Maneuver, but for the driver of the front car.
\end{enumerate}

We also consider an additional scenario consisting of the same driver maneuvers as above but for heavy vehicles, i.e., trucks and buses. In all these scenarios, for the data to resemble a real driving experience, we made use of the equipment depicted in Fig.~\ref{fig:Viena2Overal}, consisting of a steering wheel with a set of buttons and a gear stick, as well as of a set of pedals. 
We then captured images at 30 fps with a single virtual camera mounted on the vehicle and facing the road forward. Since the speed of the vehicle is displayed at a specific location in these images, we extracted it using an OCR module~\cite{smith2007overview} (see supplementary material for more detail on data collection). 
Furthermore, we developed an application that records measurements from the steering wheel. In particular, it gives us access to the steering angle every 1 microsecond, which allowed us to obtain a value of the angle synchronized with each image. Our application also lets us obtain the ground-truth label of each video sequence by recording the driver input from the steering wheel buttons. This greatly facilitated our labeling task, compared to~\cite{richterplaying,richter2016playing}, which had to use a middleware to access the rendering commands from which the ground-truth labels could be extracted. Ultimately, VIENA$^2$ consists of video sequences with synchronized measurements of steering angles and speed, and corresponding action labels.

Altogether, VIENA$^2$ contains more than 15K full HD videos (with frame size of $1920\times 1280$), corresponding to a total of more than 2.25M annotated frames. The detailed number of videos for each class and the proportions of different weather and daytime conditions of VIENA$^2$ are provided in Fig.~\ref{fig:plot_stat}. Each video contains 150 frames captured at 30 frames-per-second depicting a single action from one scenario. The action occurs in the second half of the video (mostly around the $4$ second mark), which makes VIENA$^2$ well-suited to research on action anticipation, where one typically needs to see what happens before the action starts. 

Our goal is for VIENA$^2$ to be an extensible dataset. Therefore, by making our source code and toolbox for data collection and annotation publicly available, we aim to encourage the community to participate and grow VIENA$^2$.
Furthermore, while VIENA$^2$ was mainly collected for the task of action anticipation in driving scenarios, as it contains full length videos, i.e., videos of a single drive of 30 minutes on average depicting multiple actions, it can also be used for the tasks of action recognition and temporal action localization. 

\subsection{Comparison to Other Datasets}
\label{sec:compare_datasets}

The different scenarios and action classes of VIENA$^2$ make it compatible with existing datasets, thus potentially allowing one to use our synthetic data in conjunction with real images.
For instance, the action labels in the Driver Maneuver scenario correspond to the ones in Brain4Cars~\cite{jain2016brain4cars} and in the Toyota Action Dataset~\cite{olabiyi2017driver}. Similarly, our last two scenarios dealing with heavy vehicles contain the same labels as in Brain4Cars~\cite{jain2016brain4cars}. Moreover, the actions in the Pedestrian Intention scenario corresponds to those in~\cite{ped_benchmark}.
Note, however, that, to the best of our knowledge, there is no other dataset covering our Traffic Rules and Front Car Intention scenarios, or containing data involving heavy vehicles. Similarly, there is no dataset that covers accidents involving a driver's own car. In this respect, the most closely related dataset is DashCam~\cite{chan2016anticipating}, which depicts accidents of other cars. Furthermore, VIENA$^2$ covers a much larger diversity of environmental conditions, such as daytime variations (morning, noon, afternoon, night, midnight), weather variations (clear, sunny, cloudy, foggy, hazy, rainy, snowy), and location variations (city, suburbs, highways, industrial, woods), than existing public datasets. In the supplementary material, we provide examples of each of these different environmental conditions.
In addition to covering more scenarios and conditions than other driving anticipation datasets, VIENA$^2$ also contains more samples per class than existing action analysis datasets, both for recognition and anticipation. As shown in Table~\ref{tbl:dataset_comparison}, with 600 samples per class, VIENA$^2$ outsizes (at least class-wise) the datasets that are considered \emph{large} by the community. This is also the case for other synthetic datasets, such as VIPER~\cite{richterplaying}, GTA5~\cite{richter2016playing}, VEIS~\cite{sadat2018effective}, and SYNTHIA~\cite{ros2017semantic}, which, by targeting different problems, such as semantic segmentation for which annotations are more costly to obtain, remain limited in size. We acknowledge, however, that, since we target driving scenarios, our dataset cannot match in absolute size more general recognition datasets, such as Kinetics.


\begin{table}[t]
\caption{Statistics comparison with action recognition and anticipation datasets. A * indicates a dataset specialized to one scenario, e.g., driving, as opposed to generic.}
\label{tbl:dataset_comparison}
\centering
\scriptsize
\scalebox{0.8}{
\begin{tabular}{l | c | c | c||l | c | c | c}
 & Samples &   &  &   & Samples &   & \\
Recognition & /Class & classes & videos &  Anticipation & /Class & classes & videos\\
\hline
UCF-101 (Soomro et al. 2012) & 150 & 101 & 13.3K & UT-Interaction* (Ryoo et al. 2009) & 20 & 6 & 60\\
HMDB/JHMDB (Kuehne et al. 2011) & 120 & 51/21 & 5.1K/928 & Brain4Cars* (Jain et al. 2016) & 140 & 6 & 700 \\
UCF-Sport* (Rodriguez et al. 2008) & 30 & 10 & 150  & JAAD* (Rasouli et al. 2017) & 86 & 4 & 346\\
Charades (Sigurdsson et al., 2016) & 100 & 157 & 9.8K  & & & & \\
ActivityNet (Caba et al. 2015) & 144 & 200 & 15K  & & & & \\
Kinetics (Kay et al. 2017) & 400 & 400 & 306K  & & & & \\
\hline
VIENA$^2$* & 600 & 25 & 15K & VIENA$^2$* & 600 & 25 & 15K \\
\end{tabular}
}
\end{table}

\section{Benchmark Algorithms}
\label{sec:benchmark}
In this section, we first discuss the state-of-the-art action analysis and anticipation methods that we used to benchmark our dataset. We then introduce a new multi-modal LSTM-based approach to action anticipation, and finally discuss how we model actions from our images and additional sensors.

\subsection{Baseline Methods}
The idea of anticipation was introduced in the computer vision community almost a decade ago by~\cite{ryoo2009spatio}. While the early methods
~\cite{ryoo2011human,soomro2016predicting,soomro2016online}
relied on handcrafted-features, they have now been superseded by end-to-end learning methods~\cite{ma2016learning,jain2016brain4cars,aliakbarian2017encouraging}, focusing on designing new losses better-suited to anticipation. In particular, the loss of~\cite{aliakbarian2017encouraging} has proven highly effective, achieving state-of-the-art results on several standard benchmarks. 

Despite the growing interest of the community in anticipation, action recognition still remains more thoroughly investigated. Since recognition algorithms can be converted to performing anticipation by making them predict a class label at every frame, we include the state-of-the-art recognition methods in our benchmark.
Specifically, we evaluate the following baselines:
\paragraph{Baseline 1: CNN+LSTMs.} The high performance of CNNs in image classification makes them a natural choice for video analysis, via some modifications. This was achieved in~\cite{LRCN} by feeding the frame-wise features of a CNN to an LSTM model, and taking the output of the last time-step LSTM cell as  prediction. For anticipation, we can then simply consider the prediction at each frame. We then use the temporal average pooling strategy of~\cite{aliakbarian2017encouraging}, which has proven effective to increase the robustness of the predictor for action anticipation.
\paragraph{Baseline 2: Two-Stream Networks.} 
Baseline 1 only relies on appearance, ignoring motion inherent to video (by motion, we mean explicit motion information as input, such as optical flow). Two-stream architectures, such as the one of~\cite{feichtenhofer2016convolutional}, have achieved state-of-the-art performance by explicitly accounting for motion. In particular, this is achieved by  taking a stack of 10 externally computed optical flow frames as input to the second stream. A prediction for each frame can be obtained by considering the 10 previous frames in the sequence for optical flow. We also make use of temporal average pooling of the predictions.
\paragraph{Baseline 3: Multi-Stage LSTMs.} The Multi-Stage LSTM (MS-LSTM) of~\cite{aliakbarian2017encouraging} constitutes the state of the art in action anticipation. This model jointly exploits context- and action-aware features that are used in two successive LSTM stages.
As mentioned above, the key to the success of MS-LSTM is its training loss function. This loss function can be expressed as
\begin{equation}
\mathcal{L}(y, \hat{y}) = -\frac{1}{N}\sum^N_{k=1}\sum^T_{t=1}\Bigg[y^t(k) \log(\hat{y}^t(k)) + \\ \nonumber w(t)(1-y^t(k))\log(1-\hat{y}^t(k))\Bigg]\;,
\label{eq:loss}
\end{equation}
where $y^t(k)$ is the ground-truth label of sample $k$ at frame $t$, $\hat{y}^t(k)$ the corresponding prediction, and $w(t) = \frac{t}{T}$. The first term encourages the model to predict the correct action at any time, while the second term accounts for ambiguities between different classes in the earlier part of the video.



\subsection{A New Multi-Modal LSTM}
While effective, MS-LSTM suffers from the fact that it was specifically designed to take two modalities as input, the order of which needs to be manually defined. 
As such, it does not naturally apply to our more general scenario, and must be actively modified, in what might be a sub-optimal manner, to evaluate it with our action descriptors. To overcome this, we therefore introduce a new multi-modal LSTM (MM-LSTM) architecture that generalizes the multi-stage architecture of~\cite{aliakbarian2017encouraging} to an arbitrary number of modalities. Furthermore, our MM-LSTM also aims to learn the importance of each modality for the prediction.

\begin{figure}[t]
\centering
\begin{tabular} {cc}
\includegraphics[width=.55\textwidth]{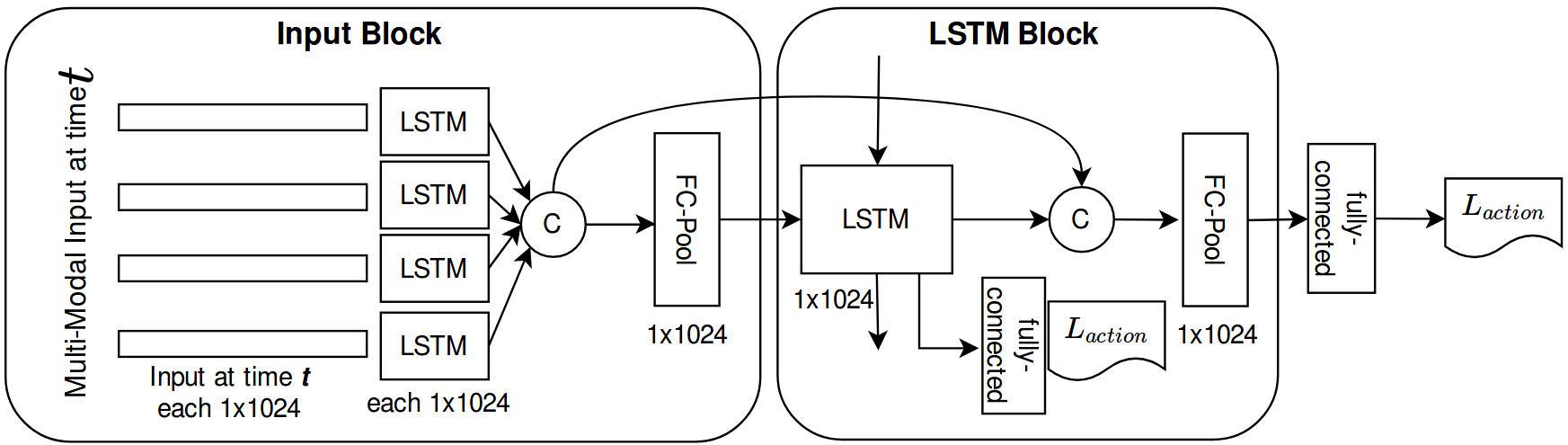} & 
\includegraphics[width=.3\textwidth]{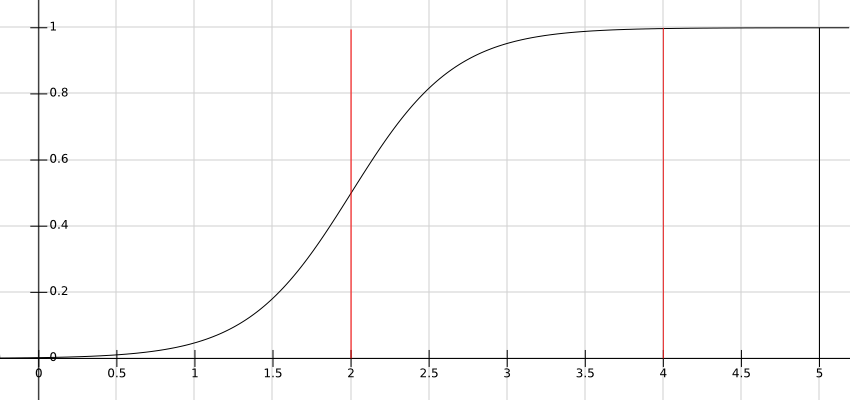} \\
Our MM-LSTM architecture & $w(t) = \frac{e^{(\alpha t-\beta)}}{1 + e^{(\alpha t-\beta)}}$\\
\end{tabular}
\caption{(Left) Our Multi-Stage LSTM architecture. (Right) Visualization of our weighting function for the anticipation loss of Eq.~\ref{eq:loss}.}
\label{fig:MM_LSTM}
\end{figure}

Specifically, as illustrated in Fig.~\ref{fig:MM_LSTM} for $M=4$ modalities, at each time $t$, the representations of the $M$ input modalities are first passed individually into an LSTM with a single hidden layer. The activations of these $M$ hidden layers are then concatenated into an $M \times 1024$ matrix $D^t$, which acts as input to a time-distributed fully-connected layer (FC-Pool). This layer then combines the $M$ modalities to form a single vector $O^t \in R^{1024}$. This representation is then passed through another LSTM whose output is concatenated with the original $D^t$ via a skip connection. The resulting $(M+1) \times 1024$ matrix is then compacted into a 1024D vector via another FC-Pool layer. The output of this FC-Pool layer constitutes the final representation and acts as input to the classification layer.

The reasoning behind this architecture is the following. The first FC-Pool layer can learn the importance of each modality. While its parameters are shared across time, the individual, modality-specific LSTMs can produce time-varying outputs, thus, together with the FC-Pool layer, providing the model with the flexibility to change the importance of each modality over time. In essence, this allows the model to learn the importance of the modalities dynamically. The second LSTM layer then models the temporal variations of the combined modalities. The skip connection and the second FC-Pool layer produce a final representation that can leverage both the individual, modality-specific representations and the learned combination of these features.

\paragraph{Learning.} To train our model, we make use of the loss of Eq.~\ref{eq:loss}. However, we modify the weights as $w(t) = \frac{e^{(\alpha t-\beta)}}{1 + e^{(\alpha t-\beta)}}$, allowing the influence of the second term to vary nonlinearly. In practice, we set $\alpha=3$ and $\beta=6$, yielding the weight function of Fig.~\ref{fig:MM_LSTM}. These values were motivated by the study of~\cite{pentland1999modeling}, which shows that driving actions typically undergo the following progression: In a first stage, the driver is not aware of an action or decides to take an action. In the next stage, the driver becomes aware of an action or decides to take one. This portion of the video contains crucial information for anticipating the upcoming action. In the last portion of the video, the action has started. In this portion of the video, we do not want to make a wrong prediction, thus penalizing false positives strongly. Generally speaking, our sigmoid-based strategy to define the weight reflects the fact that, in practice and in contrast with many academic datasets, such as UCF-101~\cite{soomro2012ucf101} and JHMDB-21~\cite{JhuangICCV2013}, actions do not start right at the beginning of a video sequence, but at any point in time, the goal being to detect them as early as possible.

During training, we rely on stage-wise supervision, by introducing an additional classification layer after the second LSTM block, as illustrated in Fig.~\ref{fig:MM_LSTM}. At test time, however, we remove this intermediate classifier to only keep the final one. We then make use of the temporal average pooling strategy of~\cite{aliakbarian2017encouraging} to accumulate the predictions over time.

\subsection{Action Modeling}
\label{sec:action_modeling}

Our MM-LSTM can take as input multiple modalities that provide diverse and complementary information about the observed data. Here, we briefly describe the different descriptors that we use in practice.

\begin{itemize}
\item {\bf Appearance-based Descriptors.} Given a frame at time $t$, the most natural source of information to predict the action is the appearance depicted in the image. To encode this information, 
we make use of a slightly modified DenseNet~\cite{huang2016densely}, pre-trained on ImageNet. See Section~\ref{sec:implementation} for more detail.
Note that we also use this DenseNet as appearance-based CNN for Baselines 1 and 2.
\item {\bf Motion-based Descriptors.} Motion has proven a useful cue for action recognition~\cite{feichtenhofer2017spatiotemporal,feichtenhofer2016convolutional}. To encode this, we make use of a similar architecture as for our appearance-based descriptors, but modify it to take as input a stack of optical flows. Specifically, we extract optical flow between $L$ consecutive pairs of frames, in the range $[t-L, t]$, and form a $2L$ flow stack encoding horizontal and vertical flows. We fine-tune the model pre-trained on ImageNet for the task of action recognition, and take the output of the additional fully-connected layer as our motion-aware descriptor. Note that we also use this DenseNet for the motion-based stream of Baseline 2. 
\item {\bf Vehicle Dynamics.} In our driving context, we have access to additional vehicle dynamics measurements. For each such measurement, at each time $t$, we compute a vector from its value $s_t$, its velocity $(s_t - s_{t-\delta})$ and its acceleration $(s_t - 2s_{t-\delta} + s_{t-2\delta})$. To map these vectors to a descriptor of size comparable to the appearance- and motion-based ones, inspired by~\cite{fernando2017going}, we train an LSTM with a single hidden layer modeling the correspondence between vehicle dynamics and action label. In our dataset, we have two types of dynamics measurements, steering angle and speed, which results in two additional descriptors. 

\end{itemize}

When evaluating the baselines, we report results of both their standard version, relying on the descriptors used in the respective papers, and of modified versions that incorporate the four descriptor types discussed above. Specifically, for CNN-LSTM, we simply concatenate the vehicle dynamics descriptors and the motion-based descriptors to the appearance-based ones. For the Two-Stream baseline, we add a second two-stream sub-network for the vehicle dynamics and merge it with the appearance and motion streams by adding a fully-connected layer that takes as input the concatenation of the representation from the original two-stream sub-network and from the vehicle dynamics two-stream sub-network. Finally, for MS-LSTM, we add a third stage that takes as input the concatenation of the second-stage representation with the vehicle dynamics descriptors. 

\subsection{Implementation Details}
\label{sec:implementation}
We make use of the DenseNet-121~\cite{huang2016densely}, pre-trained on ImageNet, to extract our appearance- and motion-based descriptors. Specifically, we replace the classifier with a fully-connected layer with 1024 neurons followed by a classifier with $N$ outputs, where $N$ is the number of classes. We fine-tune the resulting model using stochastic gradient descent for $10$ epochs with a fixed learning rate of $0.001$ and mini-batches of size $16$. Recall that, for the motion-based descriptors, the corresponding DenseNet relies on $2L$ flow stacks as input, which requires us to also replace the first layer of the network. To initialize the parameters of this layer, we average the weights over the three channels corresponding to the original RGB channels, and replicate these average weights $2L$ times~\cite{TSN}. We found this scheme to perform better than random initialization. 
\section{Benchmark Evaluation and Analysis}
We now report and analyze the results of our benchmarking experiments. For these experiments to be as extensive as possible given the available time, we performed them on a representative subset of VIENA$^2$ containing about 6.5K videos acquired in a large variety of environmental conditions and covering all 25 classes. This subset contains 277 samples per class, and thus still outsizes most action analysis datasets, as can be verified from Table~\ref{tbl:dataset_comparison}. The detailed statistics of this subset are provided in the supplementary material.

To evaluate the behavior of the algorithms in different conditions, we defined three different partitions of the data. The first one, which we refer to as \texttt{Random} in our experiments, consists of randomly assigning 70\% of the samples to the training set and the remaining 30\% to the test set. The second partition considers the daytime of the sequences, and is therefore referred to as \texttt{Daytime}. In this case, the training set is formed by the day images and the test set by the night ones. The last partition, \texttt{Weather}, follows the same strategy but based on the information about weather conditions, i.e., a training set of clear weather and a test set of rainy/snowy/... weathers.

Below, we first present the results of our benchmarking on the \texttt{Random} partition, and then analyze the challenges related to our new dataset. We finally evaluate the benefits of our synthetic data for anticipation from real images, and analyze the bias of VIENA$^2$. Note that additional results including benchmarking on the other partitions and ablation studies of our MM-LSTM model are provided in the supplementary material. Note also that the scenarios and classes acronyms are defined in Section~\ref{sec:scenarios}.

\subsection{Action Anticipation on VIENA$^2$}
We report the results of our benchmark evaluation on the different scenarios of VIENA$^2$ in Table~\ref{tbl:baseline_original} for the original versions of the baselines, relying on the descriptors used in their respective paper, and in Table~\ref{tbl:baselines_full} for their modified versions that incorporate all descriptor types. Specifically, we report the recognition accuracies for all scenarios after every second of the sequences. Note that, in general, incorporating all descriptor types improves the results. Furthermore, while the action recognition baselines perform quite well in some scenarios, such as Accidents and Traffic Rules for the two-stream model, they are clearly outperformed by the anticipation methods in the other cases. Altogether, our new MM-LSTM consistently outperforms the baselines, thus showing the benefits of learning the dynamic importance of the modalities.

\begin{table}[t]
\centering
\scriptsize
\caption{Results on the \texttt{Random} split of VIENA$^2$ for the original versions our three baselines:  CNN+LSTM~\cite{LRCN} with only appearance, Two-Stream~\cite{feichtenhofer2016convolutional} with appearance and motion, and MS-LSTM~\cite{aliakbarian2017encouraging} with action-aware and context-aware features.}
\label{tbl:baseline_original}
\scalebox{1}
{
\begin{tabular}{l| c@{ }@{ }c@{ }@{ }c@{ }@{ }c@{ }@{ }c| c@{ }@{ }c@{ }@{ }c@{ }@{ }c@{ }@{ }c| c@{ }@{ }c@{ }@{ }c@{ }@{ }c@{ }@{ }c}

&\multicolumn{5}{c}{\small CNN+LSTM~\cite{LRCN}}  & \multicolumn{5}{c}{\small Two-Stream~\cite{feichtenhofer2016convolutional}} &\multicolumn{5}{c}{\small MS-LSTM~\cite{aliakbarian2017encouraging}}\\
 \hline
 & 1" & 2" & 3" & 4" & 5" & 1" & 2" & 3" & 4" & 5" & 1" & 2" & 3" & 4" & 5"\\
\hline

DM
&   22.8     &   24.2    &   26.5    &   27.9    &   28.0     
&   23.3     &   24.8    &   30.6    &   37.5    &   41.5    
&   22.4     &   28.1    &   37.5    &   42.6   &    44.0 \\

AC
&   53.6     &   53.6    &   55.0    &   56.3    &   57.0     
&   68.5     &   70.0    &   74.5    &   76.3    &   78.0    
&   50.3     &   55.6    &   60.4    &   68.3   &    72.5 \\

TR
&    26.6    &   28.3    &   29.5    &   30.1    &   32.1     
&    28.3    &   35.6    &   44.5    &   51.5    &   53.1    
&    30.7    &   33.4    &   41.0    &   49.8   &    52.3 \\

PI
&   38.4     &   40.4    &   41.8    &   41.8    &   42.1    
&   36.8     &   37.5    &   40.0    &   40.0    &   41.2    
&   50.6     &   52.4    &   55.6    &   56.8   &    58.3 \\

FCI
&   33.0     &   36.3    &   39.5    &    39.5   &    39.6    
&   37.1     &   38.0    &   35.5    &    39.3   &    39.3   
&   44.0     &   45.3    &   51.3    &    60.2  &     63.1 \\

\end{tabular}
}
\end{table}

\begin{table}[t]
\centering
\scriptsize
\caption{Results on the \texttt{Random} split of VIENA$^2$ for our three baselines with our action descriptors and for our approach.}
\label{tbl:baselines_full}
\scalebox{0.85}
{
\begin{tabular}{l| c@{ }@{ }c@{ }@{ }c@{ }@{ }c@{ }@{ }c| c@{ }@{ }c@{ }@{ }c@{ }@{ }c@{ }@{ }c| c@{ }@{ }c@{ }@{ }c@{ }@{ }c@{ }@{ }c|   c@{ }@{ }c@{ }@{ }c@{ }@{ }c@{ }@{ }c}

&\multicolumn{5}{c}{\small CNN+LSTM~\cite{LRCN}}  & \multicolumn{5}{c}{\small Two-Stream~\cite{feichtenhofer2016convolutional}} &\multicolumn{5}{c}{\small MS-LSTM~\cite{aliakbarian2017encouraging}} &\multicolumn{5}{c}{\small Ours MM-LSTM}\\
 \hline
 & 1" & 2" & 3" & 4" & 5" & 1" & 2" & 3" & 4" & 5" & 1" & 2" & 3" & 4" & 5" & 1" & 2" & 3" & 4" & 5"\\
\hline

DM 
& 24.6 &   25.6 &   28.0 &   30.0 &  30.3  
& 26.8 & 30.5 &  40.4 &  53.4 &   62.6 
& 28.5 & 35.8 & 57.8 & 68.1 & 78.7 
& 32.0 &  38.5 &   60.5 &   71.5 &   83.6
\\

AC 
& 56.7 &  58.3 &   59.0 &  61.6 &  61.7  
& 70.0 &  72.0 &  74.0 &  77.1 &   79.7 
&  69.6 &   75.3 &   80.6 &   83.3 &   83.6 
& 76.3 &  79.0 &   81.7&   86.3 &   86.7\\

TR 
& 28.0 &  28.7 &  30.6 &  32.2 &  32.8  
& 30.6 &  38.7 &  48.0 & 49.6 &  54.1 
&  33.3 & 39.4 & 48.3 & 57.1 & 61.0 
& 39.8 &   49.8 &   58.8 &   63.7 &   68.8\\

PI
& 39.6 & 39.6 & 40.4 & 42.0 & 42.4
& 42.0 &  42.8 &  44.4 &  46.0 &  48.0 
&  55.8 & 57.6 & 62.6 &69.0 & 70.8 
& 57.3 &   59.7 &   68.9 &   72.5 &   73.3\\

FCI 
& 37.2 & 38.8 & 39.3 & 40.6 & 40.6
& 37.7 &  39.1 &  39.3 &  40.7 &   43.0
& 41.7 & 49.1 & 58.3 & 70.0 & 75.5
& 49.9 &   51.7 &  60.4 &   71.5 &   77.8
\end{tabular}
}
\end{table}

A comparison of the baselines with our approach on the \texttt{Daytime} and \texttt{Weather} partitions of VIENA$^2$ is provided in the supplementary material. In essence, the conclusions of these experiments are the same as those drawn above.
\subsection{Challenges of VIENA$^2$}
Based on the results above, we now study what challenges our dataset brings, such as which classes are the most difficult to predict and which classes cause the most confusion. We base this analysis on the per-class accuracies of our MM-LSTM model, which achieved the best performance in our benchmark. This, we believe, can suggest new directions to investigate in the future.

\begin{figure*}[t]
\centering
\scriptsize
\begin{tabular}{c @{ }@{ }@{ }@{ } c @{ }@{ }@{ }@{ } c @{ }@{ }@{ }@{ } c @{ }@{ }@{ }@{ }c}
\scriptsize{DM}   & \scriptsize{AC} & \scriptsize{TR}  & \scriptsize{PI}  & \scriptsize{FCI}  \\
\includegraphics[width=.13\textwidth]{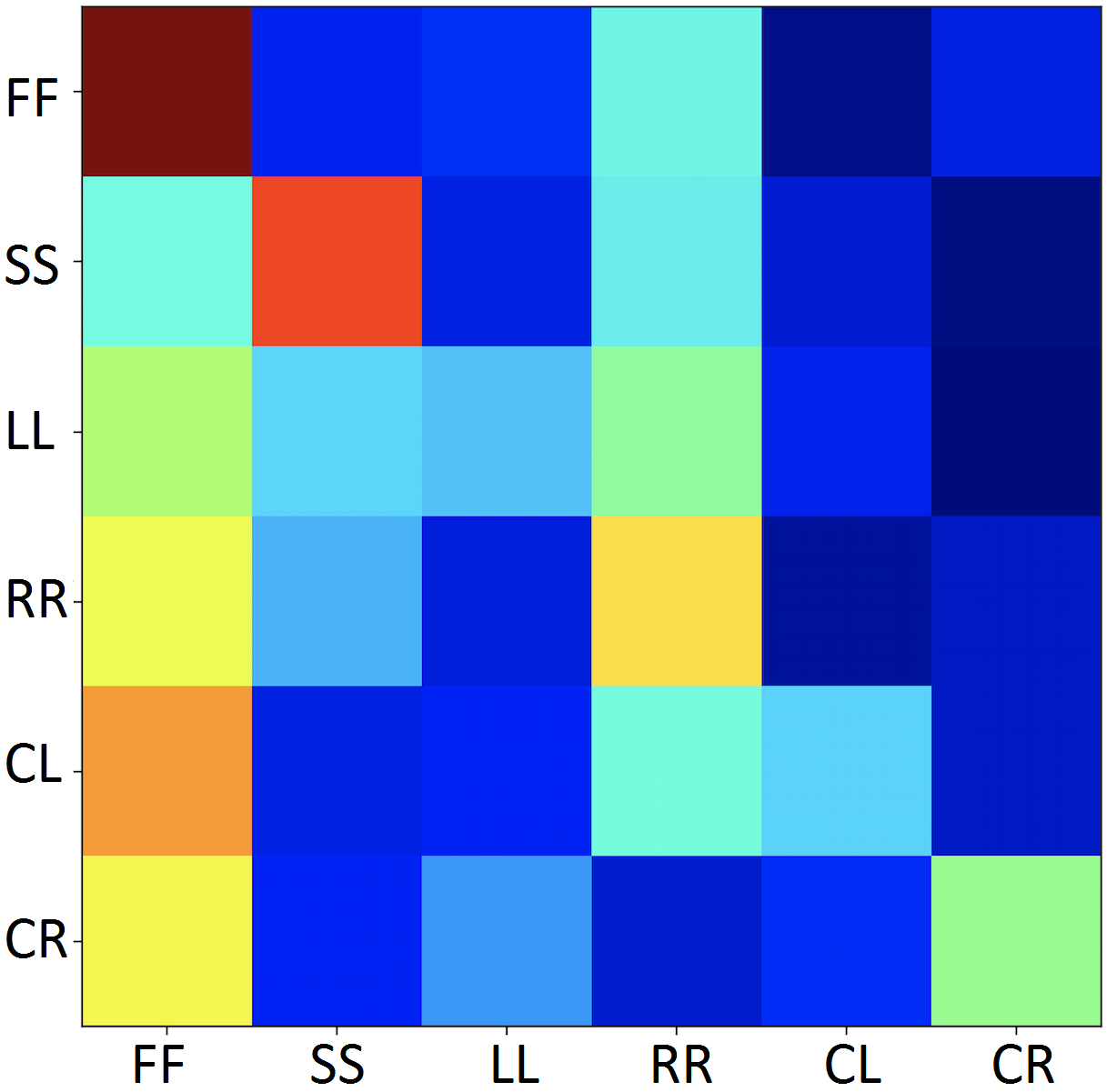} & 
\includegraphics[width=.13\textwidth]{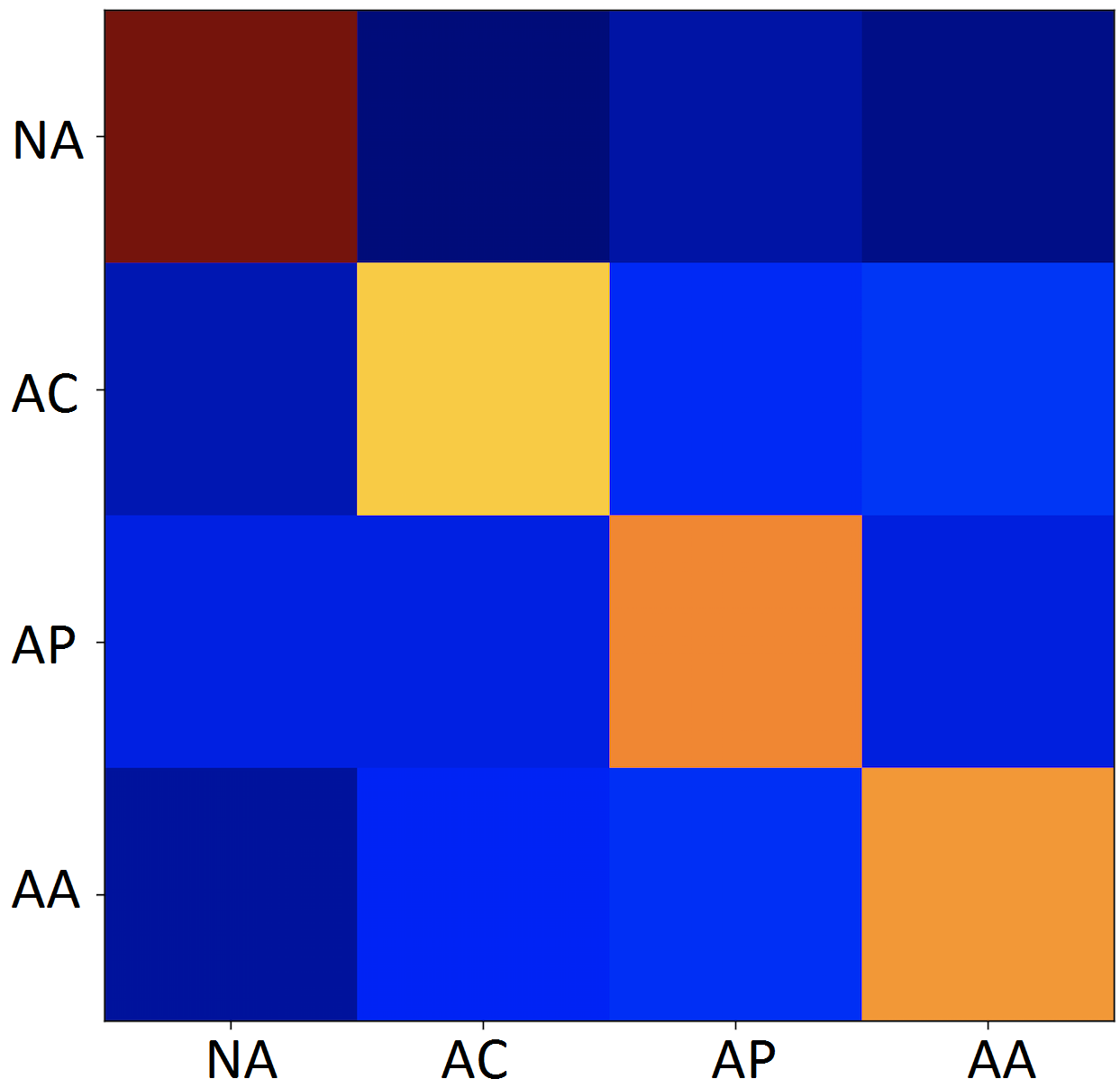} & 
\includegraphics[width=.13\textwidth]{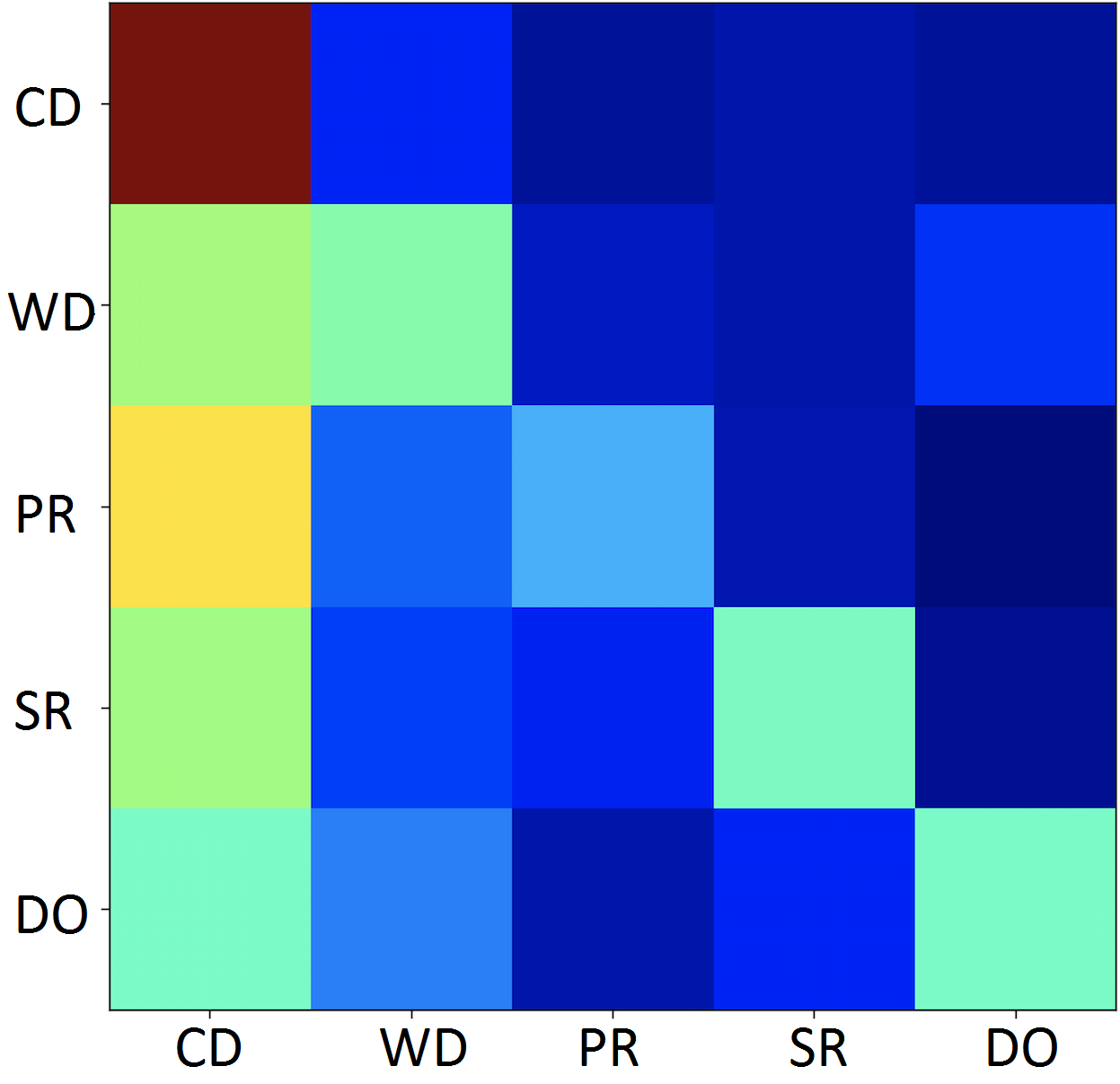} & 
\includegraphics[width=.13\textwidth]{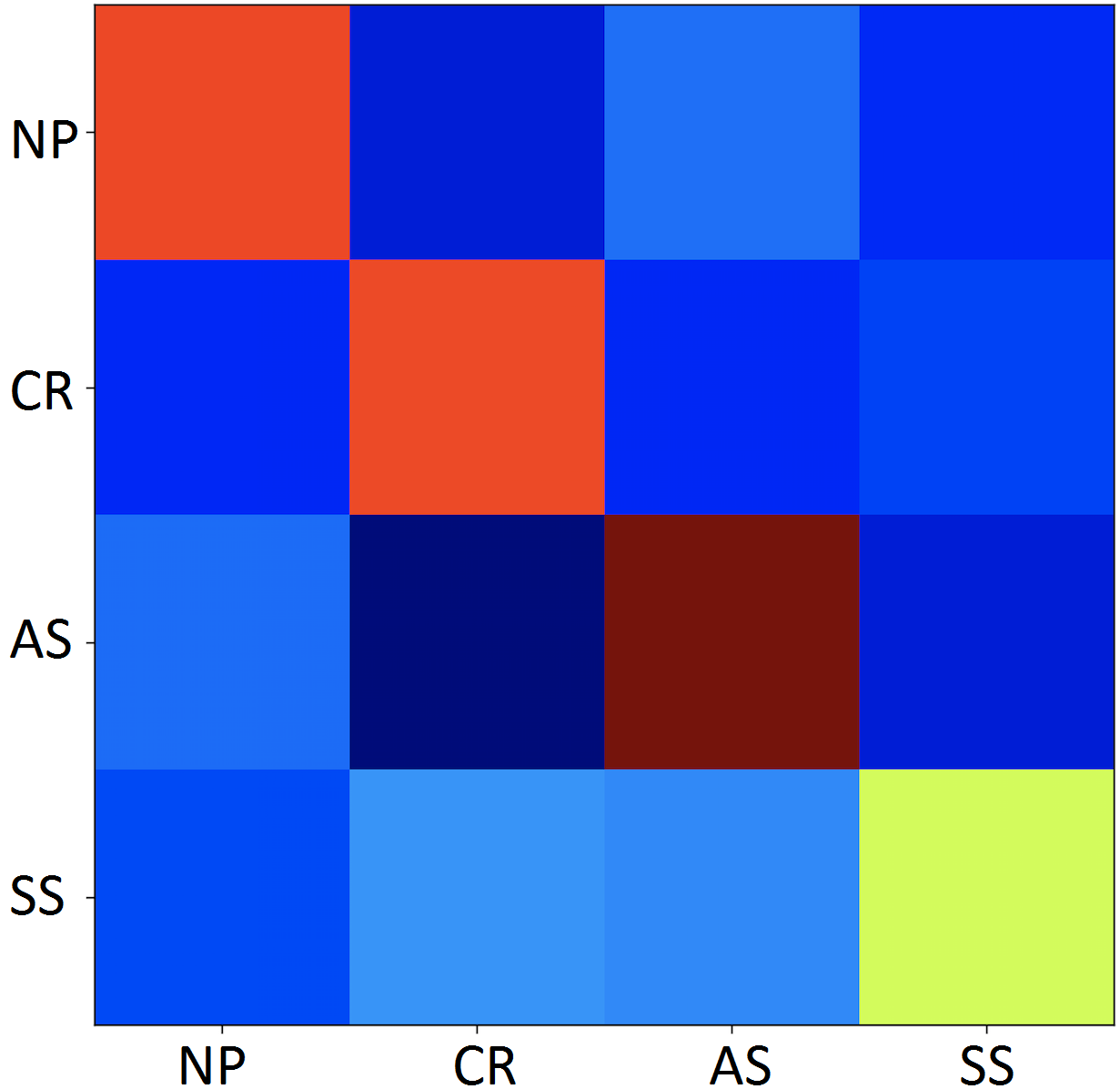} & 
\includegraphics[width=.13\textwidth]{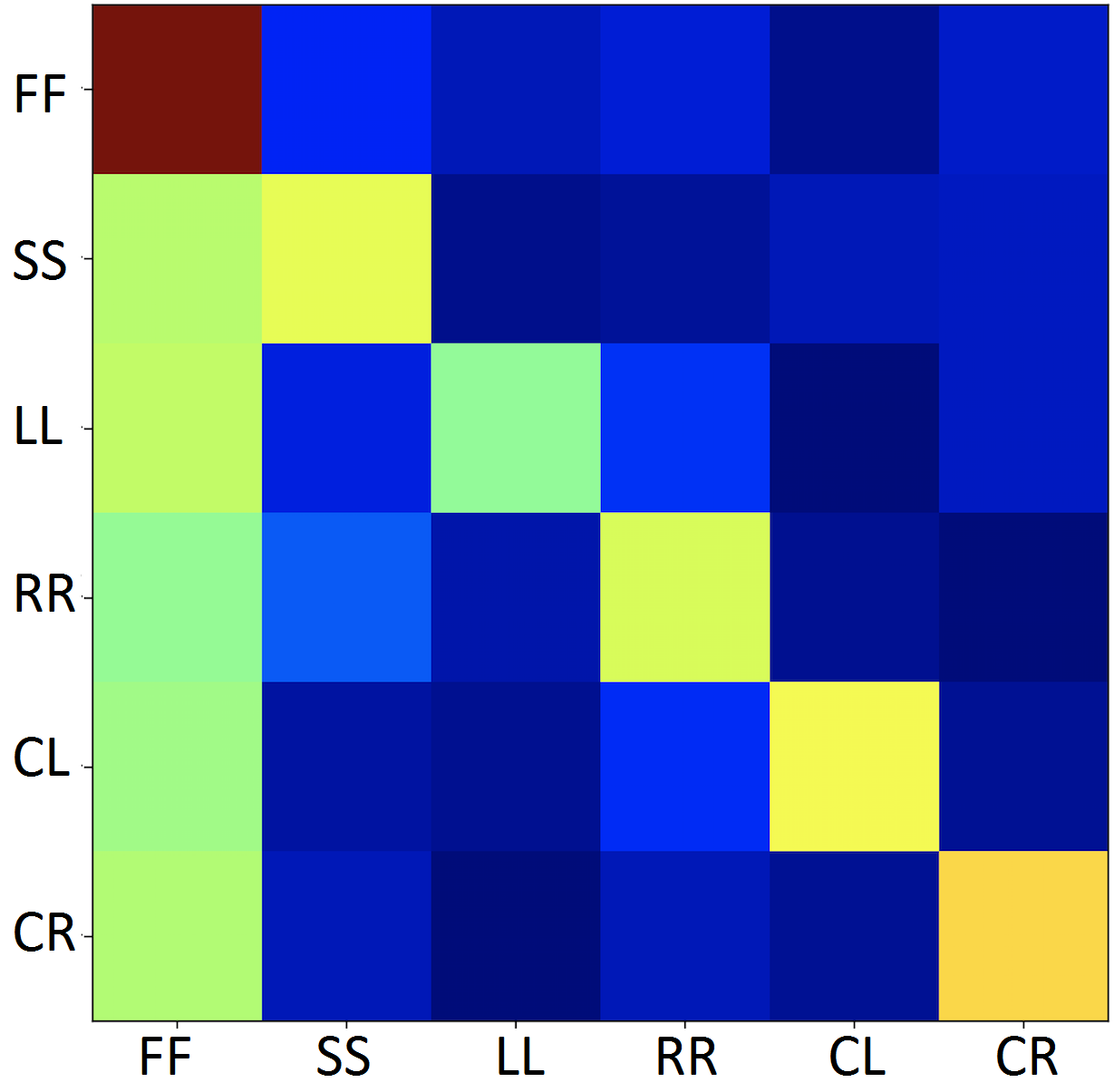} \\
\multicolumn{5}{c}{Confusion matrices after \textbf{1} second.}\\
\includegraphics[width=.13\textwidth]{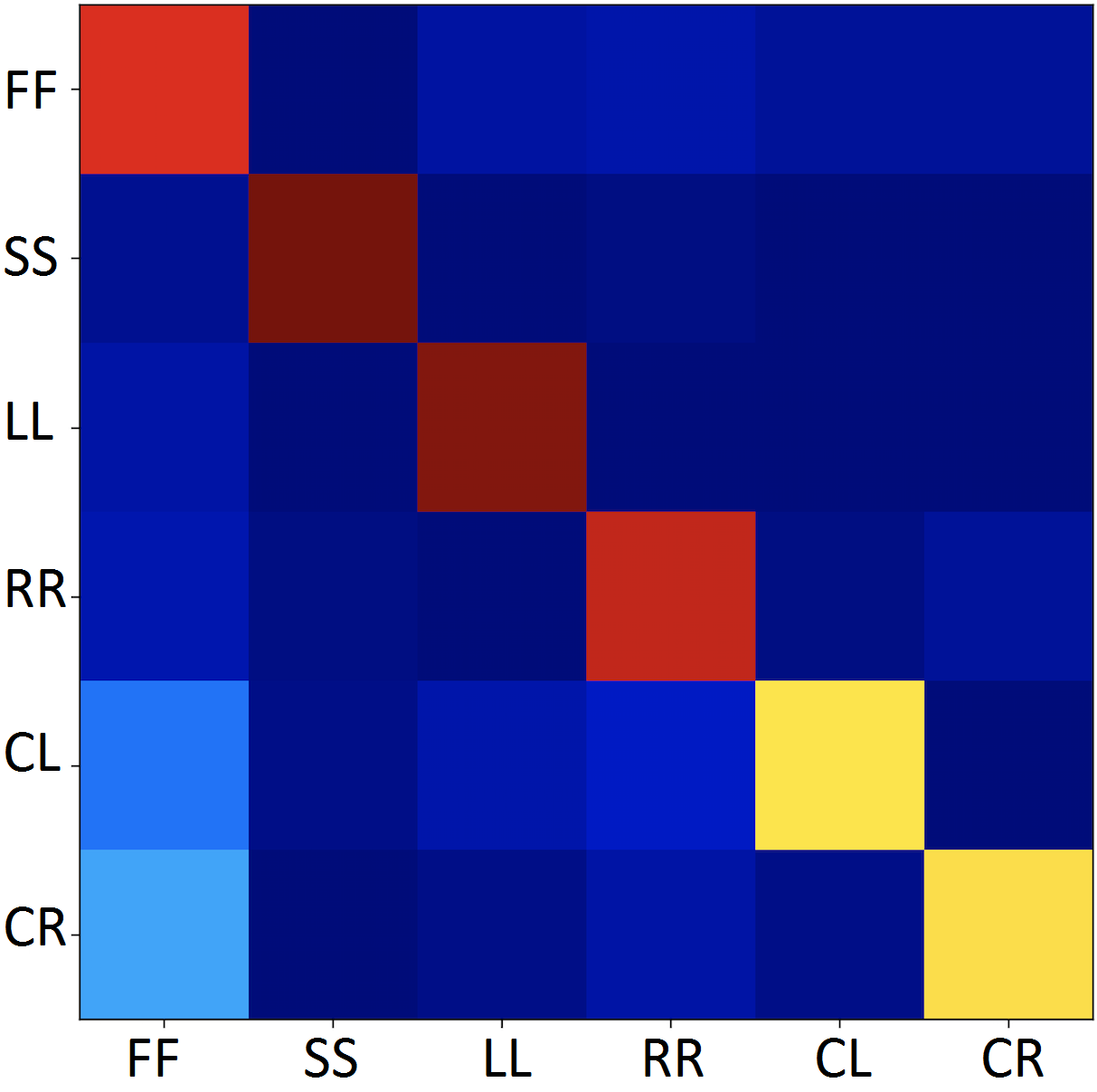} & 
\includegraphics[width=.13\textwidth]{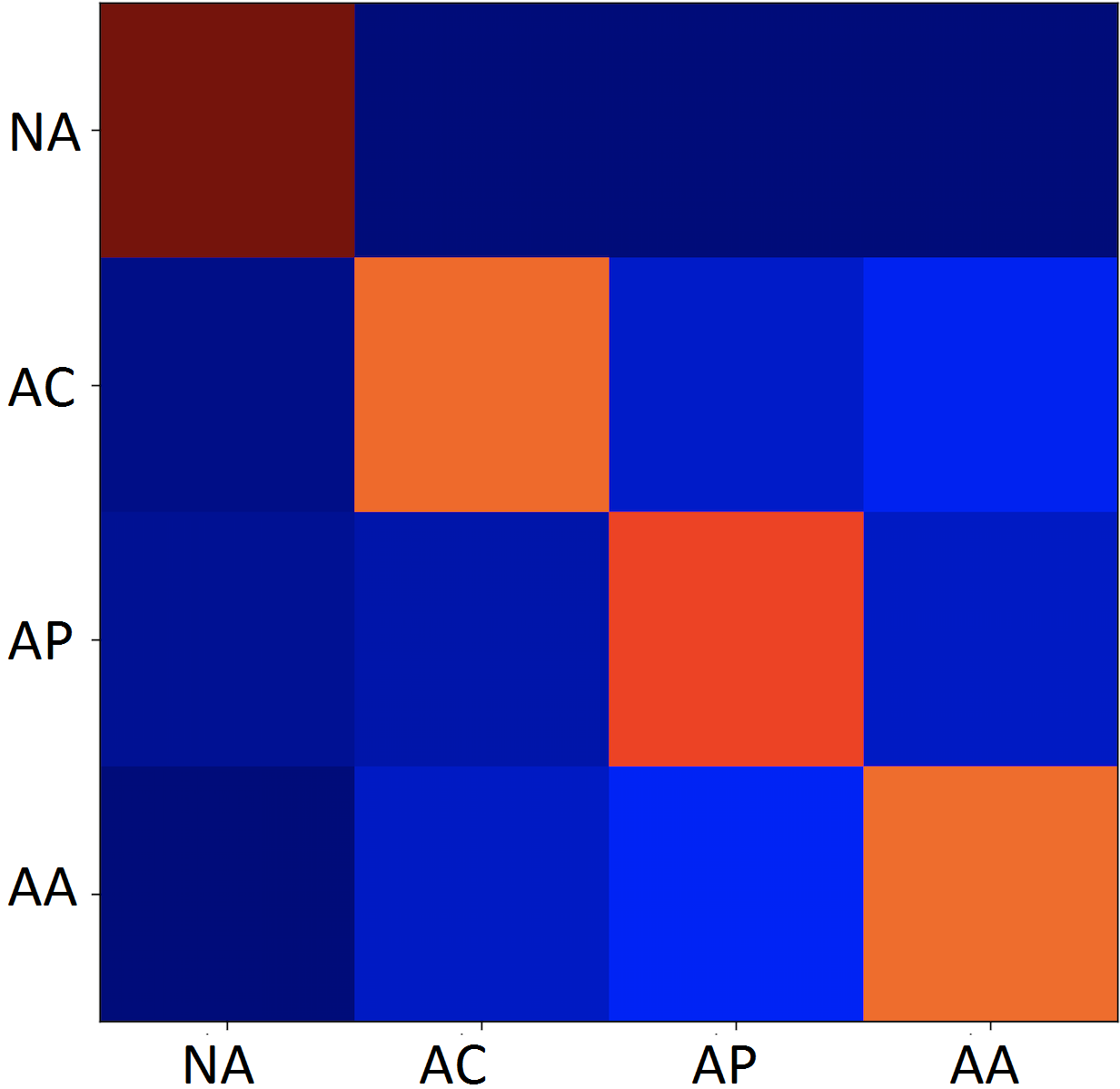} & 
\includegraphics[width=.13\textwidth]{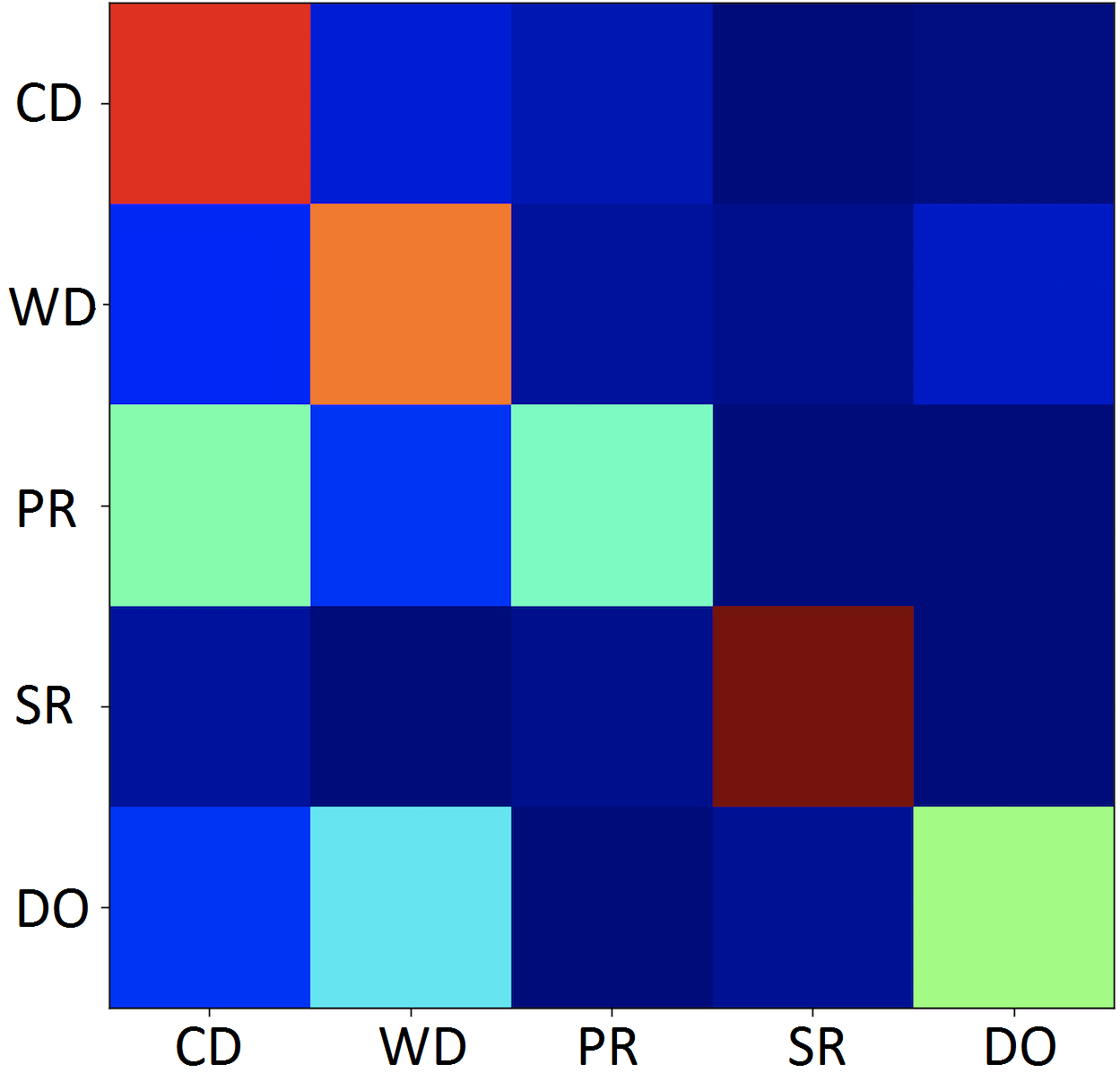} & 
\includegraphics[width=.13\textwidth]{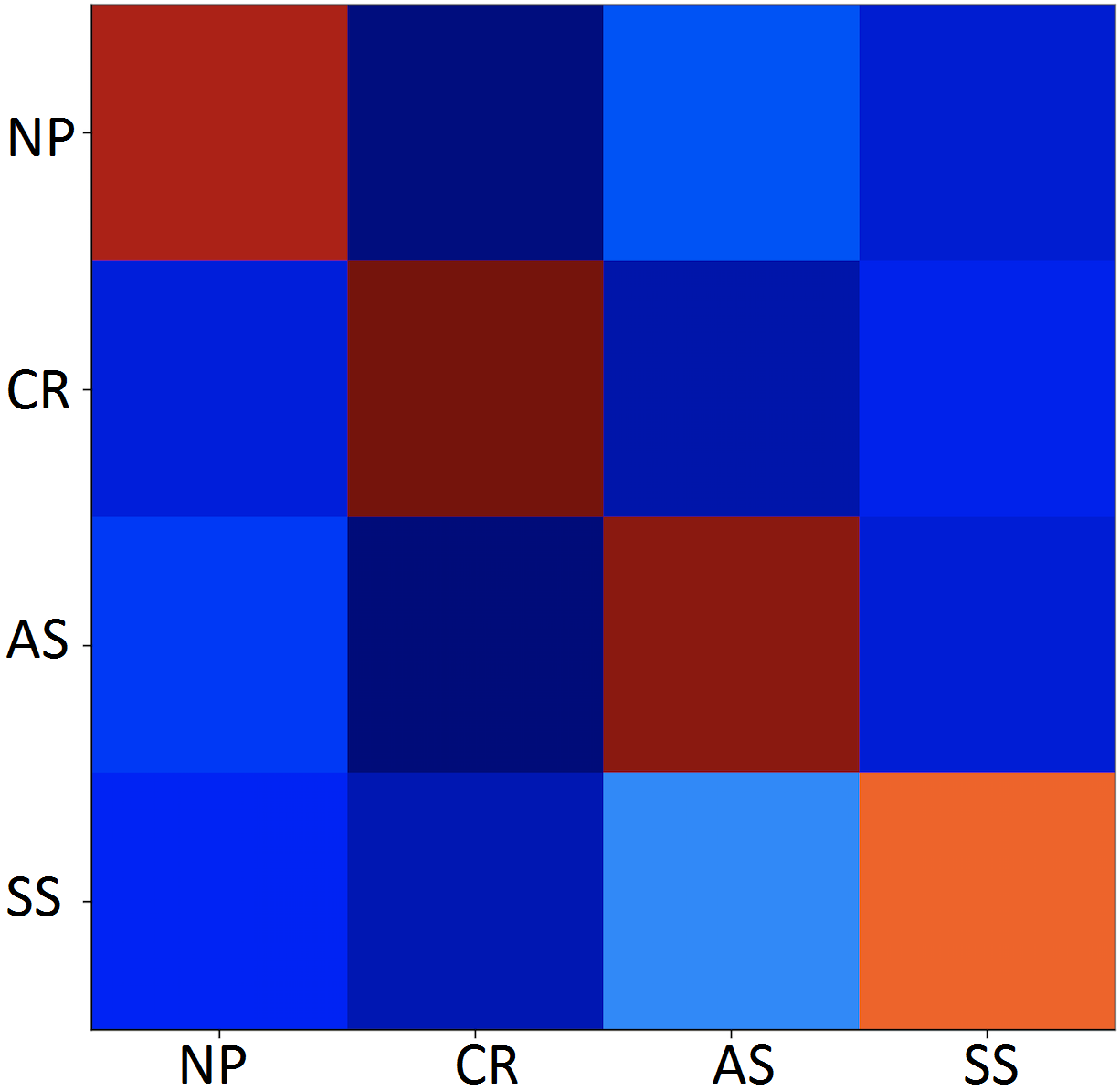} & 
\includegraphics[width=.13\textwidth]{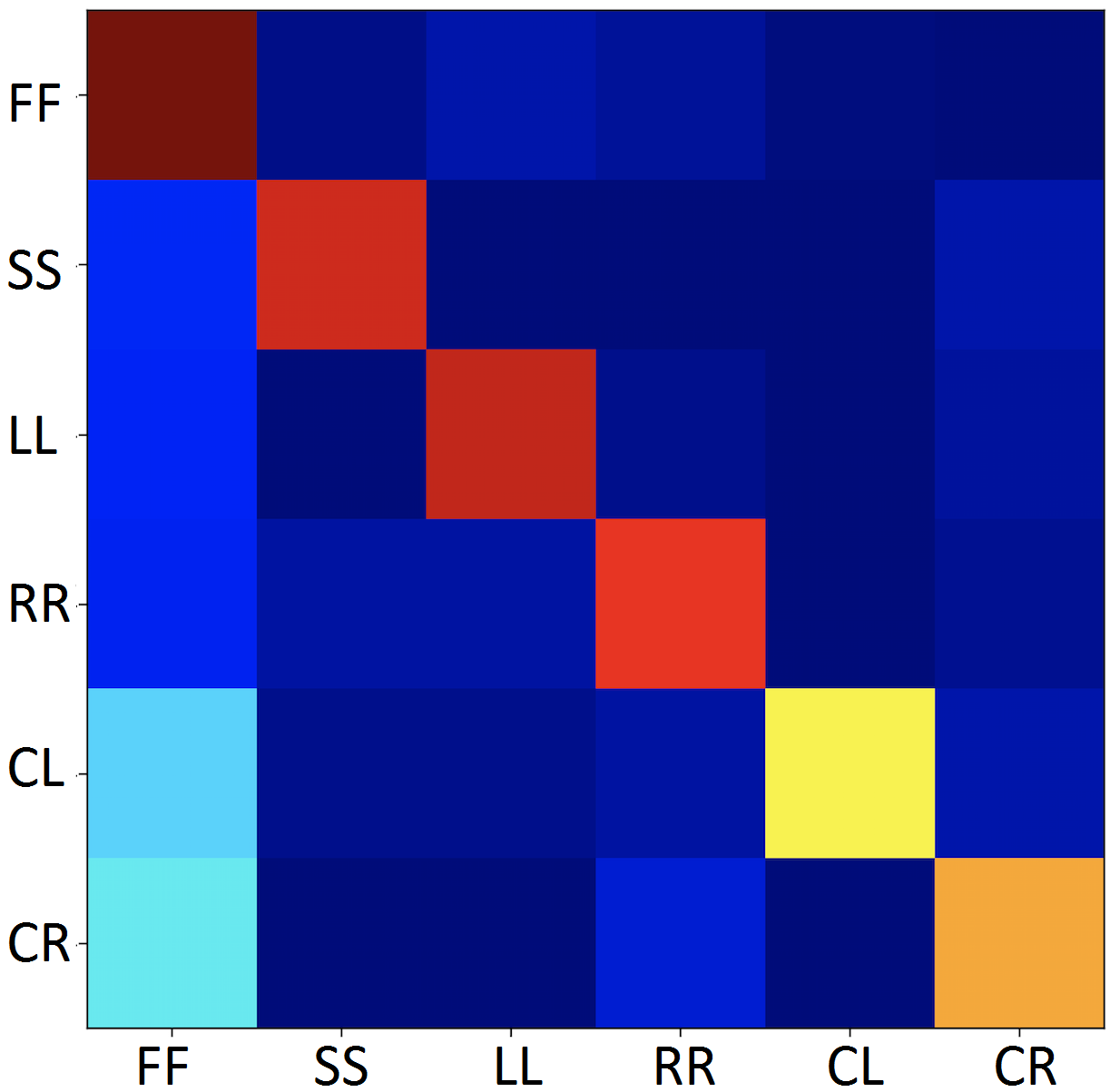} \\
\multicolumn{5}{c}{Confusion matrices after \textbf{5} second.}\\
\end{tabular}
\caption{\textbf{Confusion Matrices.} Confusion matrices of all five scenarios after observing 1 second (top) and 5 seconds (bottom) of each video sample.}
\label{fig:confusion}
\end{figure*}

Our MM-LSTM per-class accuracies are provided in Table~\ref{tbl:per_class}, and the corresponding confusion matrices at the earliest (after seeing 1 second) and latest (after seeing 5 seconds) predictions in Fig.~\ref{fig:confusion}. Below, we discuss the challenges of the various scenarios. 

\begin{table}[t]
\renewcommand{\arraystretch}{1.2}
\centering
\caption{Per-class accuracy of our approach on all scenarios of VIENA$^2$ (\texttt{Random}).}
\label{tbl:per_class}
\tiny
\scalebox{0.88}
{
\begin{tabular}{l | c c c c c c | c c c c | c c c c c | c c c c | c c c c c c}

&\multicolumn{6}{c}{\small DM} & \multicolumn{4}{c}{\small AC} & \multicolumn{5}{c}{\small TR} & \multicolumn{4}{c}{\small PI} & \multicolumn{6}{c}{\small FCI} \\
\hline
&{\scriptsize FF} & {\scriptsize SS} & {\scriptsize LL} & {\scriptsize RR} & {\scriptsize CL} & {\scriptsize CR}  
 & {\scriptsize NA} & {\scriptsize AP} & {\scriptsize AC} & {\scriptsize AA}
 & {\scriptsize CD} & {\scriptsize WD} & {\scriptsize PR} & {\scriptsize SR} & {\scriptsize DO}
 & {\scriptsize NP} & {\scriptsize CR} & {\scriptsize SS} & {\scriptsize AS}
 & {\scriptsize FF} & {\scriptsize SS} & {\scriptsize LL} & {\scriptsize RR} & {\scriptsize CL} & {\scriptsize CR}    \\
 \hline
1" &  50.7 &   43.8 &   17.8 &   35.0 &  18.7  &  26.1  
& 94.9 &   65.7 &   73.2 &   71.3
& 75.5 &   35.0 &   23.7 &   32.8 &  32.2 
& 59.3 &   59.1 &   68.8 &   42.2
& 74.5 &   46.3 &   35.6 &   44.6 &  47.8  & 50.9  
\\ 
2" &   60.1 &   46.8 &   26.3 &   38.7 &  27.1  &  32.1 
& 98.7 &   70.7 &   71.6 &   75.0
& 79.6 &   49.3 &   29.7 &  52.3  &  37.9 
& 63.0 &   51.2 &  71.4 &  53.4
& 76.9 &   48.6 &   37.1 &   45.9 &  49.6  &  52.0  
\\ 
3" &   81.3 &   75.6 &   54.4 &   63.4 &  42.9  &  45.4  
& 100 &   75.4 &   76.1 &   75.2
& 83.7 &   60.0 &  35.1 &  69.5 &  45.8 
& 70.4 &   67.6 &  79.2 &   58.6
& 85.7 &   63.9 &   54.1 &   50.7 &  52.7  &  57.3  
\\ 
4" &   81.2 &   87.3 &   72.9 &   77.3 &  55.4  &  55.0  
& 100 &   81.6 &   79.4 &   84.3
& 86.7 &  65.3 &   37.9 &  78.7 &  50.0 
& 72.2 &   75.9 &  80.1 &  61.35
& 89.1 &  77.8 &   74.4 &  69.5 & 56.1 & 62.2  
\\ 
5" &   88.0 &   97.2 &   95.8 &   90.4 &  64.9  &  65.4  
& 100 &   80.5 &   86.1 &   80.2
& 85.7 &   75.0 &  40.0 &  95.1 &  48.6 
& 74.1 &   78.2 &   76.6 &  63.6
& 91.2 &  83.5 &  84.6 &  81.4 & 59.4  &  66.8  
\\ 
\end{tabular}
}
\end{table}

\begin{enumerate}
\item {\bf Driver maneuver:} After 1s, most actions are mistaken for \emph{Moving Forward}, which is not surprising since the action has not started yet. After 5s, most of the confusion has disappeared, except for \emph{Changing Lane} (left and right), for which the appearance, motion and vehicle dynamics are subject to small changes only, thus making this action look similar to \emph{Moving Forward}.
\item {\bf Accident:} Our model is able to distinguish \emph{No Accident} from the different accident types early in the sequence. Some confusion between the different types of accident remains until after 5s, but this would have less impact in practice, as long as an accident is predicted.
\item {\bf Traffic rule:} As in the maneuver case, there is initially a high confusion with \emph{Correct Direction}, due to the fact that the action has not started yet. The confusion is then much reduced as we see more information, but \emph{Passing a Red Light} remains relatively poorly predicted.
\item {\bf Pedestrian intention:} The most challenging class for early prediction in this scenario is \emph{Pedestrian Walking along the Road}. The prediction is nevertheless much improved after 5s.
\item {\bf Front car intention:} Once again, at the beginning of the sequence, there is much confusion with the \emph{Forward} class. After 5s, the confusion is significantly reduced, with, as in the maneuver case, some confusion remaining between the \emph{Change lane} classes and the \emph{Forward} class, illustrating the subtle differences between these actions.
\end{enumerate}
\subsection{Benefits of VIENA$^2$ for Anticipation from Real Images}
\label{sec:real_images}
To evaluate the benefits of our synthetic dataset for anticipation on real videos, we make use of the JAAD dataset~\cite{rasouli2017agreeing} for pedestrian intention recognition, which is better suited to deep networks than other datasets, such as~\cite{ped_benchmark}, because of its larger size (58 videos vs. 346). This dataset is, however, not annotated with the same classes as we have in VIENA$^2$, as its purpose is to study pedestrian and driver behaviors at pedestrian crossings. To make JAAD suitable for our task, we re-annotated its videos according to the four classes of our \emph{Pedestrian Intention} scenario, and prepared a corresponding train/test split. JAAD is also heavily dominated by the \emph{Crossing} label, requiring augmentation of both training and test sets to have a more balanced number of samples per class. 

To demonstrate the benefits of  VIENA$^2$ in real-world applications, we conduct two sets of experiments: 1) Training on JAAD from scratch, and 2) Pre-training on VIENA$^2$ followed by fine-tuning on JAAD.
For all experiments, we use appearance-based and motion-based features, which can easily be obtained for JAAD. The results are shown in Table~\ref{tbl:jaad}. This experiment clearly demonstrates the effectiveness of using our synthetic dataset that contains photo-realistic samples simulating real-world scenarios.

\begin{table}[t]
\centering
\scriptsize
\caption{\textbf{Anticipating actions on real data.} Pre-training our MM-LSTM with our VIENA$^2$ dataset yields higher accuracy than training from scratch on real data.
}
\label{tbl:jaad}
\begin{tabular}{l| @{ }@{ }@{ } c @{ }@{ }@{ } c @{ }@{ }@{ } c @{ }@{ }@{ } c @{ }@{ }@{ } c}
Setup & After 1" & After 2" & After 3" & After 4" & After 5" \\
\hline
From Scratch & 41.01\% & 45.84\% & 51.38\% & 54.94\% & 56.12\% \\
Fine-Tuned & 45.06\% & 54.15\% & 58.10\% & 65.61\% & 66.0\% \\
\end{tabular}
\end{table}

\begin{figure}[t]
\centering
\includegraphics[width=.6\textwidth]{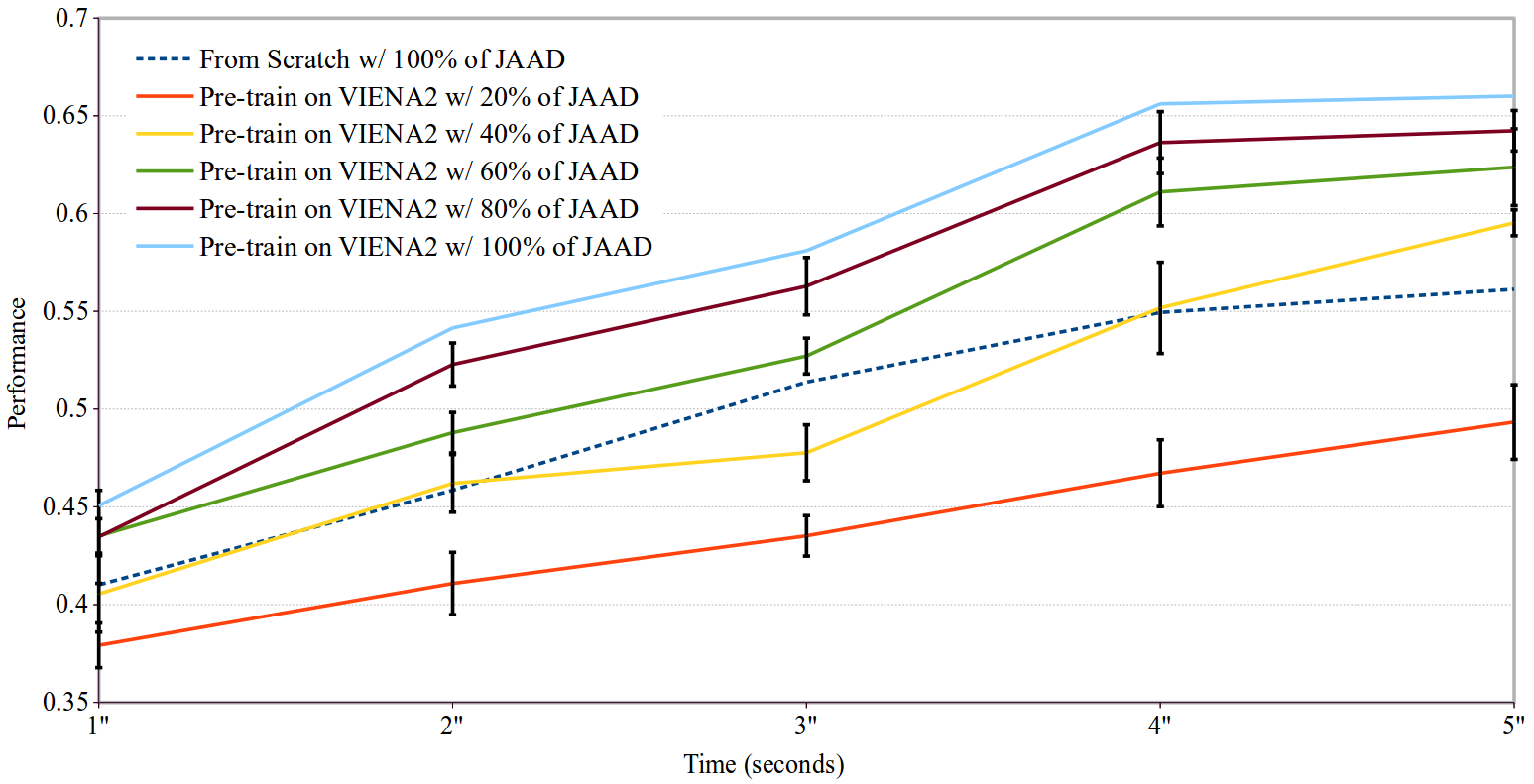}
\caption{{\bf Effect of the amount of real training data for fine-tuning MM-LSTM.} MM-LSTM was pre-trained on VIENA$^2$ in all cases, except for \emph{From Scratch w/100\% of JAAD} (dashed line). Each experiment was conducted with 10 random subsets of JAAD. We report the mean accuracy and standard deviation (error bars) over 10 runs.}
\label{fig:jaad_perc}
\end{figure}

Another potential benefit of using synthetic data is that it can reduce the amount of real data required to train a model. To evaluate this, we fine-tuned an MM-LSTM trained on VIENA$^2$ using a random subset of JAAD ranging from 20\% to 100\% of the entire dataset. The accuracies at every second of the sequence and for different percentages of JAAD data are shown in Fig.~\ref{fig:jaad_perc}. Note that with 60\% of real data, our MM-LSTM pre-trained on VIENA$^2$ already outperforms a model trained from scratch on 100\% of the JAAD data. This shows that our synthetic data can save a considerable amount of labeling effort on real images. 

\subsection{Bias Analysis}

For a dataset to be unbiased, it needs to be representative of the entire application domain it covers, thus being helpful in the presence of other data from the same application domain. This is what we aimed to achieve when capturing data in a large diversity of environmental conditions. Nevertheless, every dataset is subject to some bias. For example, since our data is synthetic, its appearance differs to some degree from real images, and the environments we cover are limited by those of the GTA V video game. However, below, we show empirically that the bias in VIENA$^2$ remains manageable, making it useful beyond evaluation on VIENA$^2$ itself.
In fact, the experiments of Section~\ref{sec:real_images} on real data already showed that performance on other datasets, such as JAAD, can be improved by making use of VIENA$^2$. To further evaluate the bias of the visual appearance of our dataset, we relied on the idea of domain adversarial training introduced in~\cite{ganin2015unsupervised}. In short, given data from two different domains, synthetic and real in our case, domain adversarial training aims to learn a feature extractor, such as a DenseNet, so as to fool a classifier whose goal is to determine from which domain a sample comes. If the visual appearance of both domains is similar, such a classifier should perform poorly. We therefore trained a DenseNet to perform action classification from a single image using both VIENA$^2$ and JAAD data, while learning a domain classifier to discriminate real samples from synthetic ones. The performance of the domain classifier quickly dropped down to chance, i.e., 50\%. To make sure that this was not simply due to failure to effectively train the domain classifier, we then froze the parameters of the DenseNet while continuing to train the domain classifier. Its accuracy remained close to chance, thus showing that the features extracted from both domains were virtually indistinguishable. Note that the accuracy of action classification improved from 18\% to 43\% during the training, thus showing that, while the features are indistinguishable to the discriminator, they are useful for action classification.

\begin{table}[t]
\centering
\scriptsize
\caption{ Effect of data collector on MM-LSTM performance (DM scenario).}
\label{tab:new_driver}
\begin{tabular}{l@{ }@{ }@{ } l@{ }@{ }  | @{ }c @{ }@{ }@{ }c @{ }@{ }@{ }c @{ }@{ }@{ }c @{ }@{ }@{ }c}
Train, captured by & Test, captured by & After 1" & After 2" & After 3" & After 4" & After 5" \\
\hline
User 1 & User 1 & 32.0\% &  38.5\% &   60.5\% &   71.5\% &   83.6\% \\
User 1 & User 2 & 32.8\% & 37.3\% & 60.7\% & 70.9\% & 82.8\% \\
\end{tabular}
\end{table}

In our context of synthetic data, another source of bias could arise from the specific users who captured the data. To analyze this, we trained an MM-LSTM model from the data acquired by a single user, covering all classes and all environmental conditions, and tested it on the data acquired by another user. In Table~\ref{tab:new_driver}, we compare the average accuracies of this experiment to those obtained when training and testing on data from the same user. Note that there is no significant differences, showing that our data generalizes well to other users.

\section{Conclusion} 
\label{sec:conclusion}
We have introduced a new large-scale dataset for general action anticipation in driving scenarios, which covers a broad range of situations with a common set of sensors. Furthermore, we have proposed a new MM-LSTM architecture allowing us to learn the importance of multiple input modalities for action anticipation. Our experimental evaluation has shown the benefits of our new dataset and of our new model.
Nevertheless, much progress remains to be done to make anticipation reliable enough for automated driving. In the future, we will therefore investigate the use of additional descriptors and of dense connections within our MM-LSTM architecture. We will also extend our dataset with more scenarios and other types of vehicles, such as motorbikes and bicycles, whose riders are more vulnerable road users than drivers. Moreover, we will extend our annotations so that every frame is annotated with bounding boxes around critical objects, such as pedestrians, cars, and traffic lights.


\bibliographystyle{splncs04}
\bibliography{0353}

\end{document}